\pdfoutput=1

\PassOptionsToPackage{table}{xcolor}

\documentclass[11pt]{article}

\usepackage[preprint]{acl}

\usepackage{times}
\usepackage{latexsym}

\usepackage[T1]{fontenc}

\usepackage[utf8]{inputenc}

\usepackage{microtype}

\usepackage{inconsolata}
\usepackage{adjustbox}
\usepackage{enumitem}
\setitemize{noitemsep,topsep=0pt,parsep=0pt,partopsep=0pt}
\definecolor{cite}{rgb}{0.6,0.6,1.0}

\definecolor{todo}{rgb}{1,0.5,0}

\usepackage{booktabs}
\usepackage{graphicx}
\usepackage{float}
\usepackage{tabularx}
\usepackage{soul}
\usepackage{subcaption}
\usepackage{multicol}
\usepackage{multirow}
\usepackage{amsfonts}
\usepackage{tabularx}
\usepackage{makecell}

\def\benchmarkname{LAB}
\usepackage{amsmath}
\usepackage{pdflscape}
\usepackage{afterpage}
\usepackage{capt-of}
\usepackage[graphicx]{realboxes}
\usepackage{rotating}
\definecolor{blue}{RGB}{218, 232, 252}
\definecolor{green}{RGB}{213, 232, 212}
\definecolor{red}{RGB}{248, 206, 215}
\definecolor{diffred}{RGB}{156, 2, 2}
\definecolor{diffgreen}{RGB}{2, 156, 2}

\title{Attribute or Abstain: Large Language Models as Long Document Assistants}

\author{Jan Buchmann, Xiao Liu, Iryna Gurevych \\
  Ubiquitous Knowledge Processing (UKP) Lab \\
  Department of Computer Science and Hessian Center for AI (hessian.AI) \\
  Technical University of Darmstadt \\
  \url{www.ukp.tu-darmstadt.de}}

\begin{document}
\maketitle
\begin{abstract}
LLMs can help humans working with long documents, but are known to hallucinate. \textit{Attribution} can increase trust in LLM responses: The LLM provides evidence that supports its response, which enhances verifiability. Existing approaches to attribution have only been evaluated in RAG settings, where the initial retrieval confounds LLM performance. This is crucially different from the long document setting, where retrieval is not needed, but could help. Thus, a long document specific evaluation of attribution is missing. To fill this gap, we present \benchmarkname{}, a benchmark of 6 diverse long document tasks with attribution, and experiments with different approaches to attribution on 5 LLMs of different sizes. \\
We find that \texttt{citation}, i.e. response generation and evidence extraction in one step, performs best for large and fine-tuned models, while additional retrieval can help for small, prompted models. We investigate whether the ``Lost in the Middle'' phenomenon exists for attribution, but do not find this. We also find that evidence quality can predict response quality on datasets with simple responses, but not so for complex responses, as models struggle with providing evidence for complex claims. We release code and data for further investigation\footnote{\href{https://github.com/UKPLab/arxiv2024-attribute-or-abstain}{Github repository}, code under Apache 2.0, dataset licenses depend on original license, see §\ref{sec:appendix/datasets}.}.
\end{abstract}

\section{Introduction}
\label{sec:introduction}
 \begin{figure}
    \centering
    \includegraphics{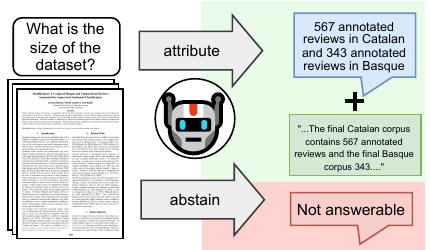}
    \caption{Long document assistants should \textit{attribute}, i.e. provide \colorbox{blue}{responses} with \colorbox{green}{evidence}, or \colorbox{red}{abstain}. Example from QASPER \cite{dasigi-etal-2021-dataset}. Figure requires emojis to display correctly.}
    \label{fig:eyecatcher}
\end{figure}

\begin{figure*}
    \centering
    \includegraphics[scale=0.95]{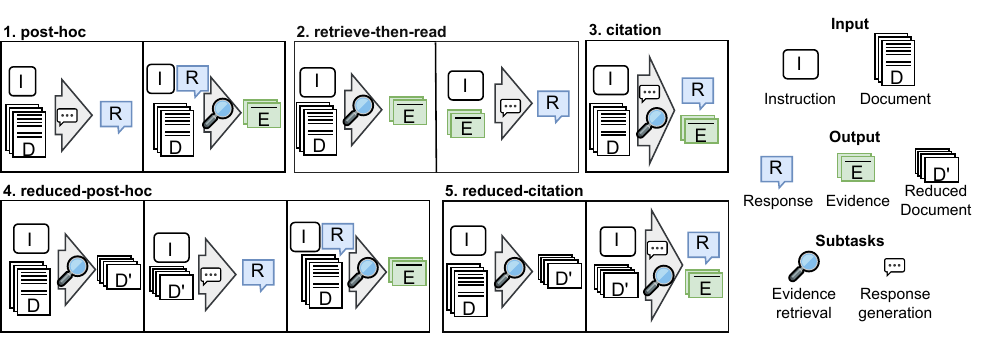}
    \caption{The approaches to attribution in long document scenarios analyzed in this work.}
    \label{fig:attribution}
\end{figure*}

Recent LLMs can process long documents \cite{shaham-etal-2023-zeroscrolls, li2023loogle}, showing great potential as \textit{long document assistants}. For example (Fig.~\ref{fig:eyecatcher}), such an assistant could answer a researcher's questions about a paper. However, due to LLM hallucinations \cite{slobodkin2023curious}, the researcher must verify responses, which is difficult with lengthy papers. To improve verifiability and trust, the assistant should either \textit{attribute} \cite{rashkin2023attribution} or \textit{abstain} \cite{slobodkin2023curious}: If it finds the necessary information, it should provide a response and point to the evidence in the paper (attribute). If not, it should clearly communicate this (abstain). We investigate the capabilities of LLMs to fulfill these requirements, and the relation between \textit{response quality} (i.e. correctness) and \textit{evidence quality} (i.e. the relevance of the evidence to the response).

Thus far, attribution has only been investigated in retrieval augmented generation (RAG) settings, where evidence comes from a large corpus (e.g. Wikipedia) that does not fit the LLM context. This means that some form of retrieval is required to provide the LLM with a limited number of relevant passages or to retrieve evidence \textit{post-hoc}. Performance depends on retrieval quality, and the best-performing approach is unclear \cite{bohnet2022attributed, gao2023alce, malaviya2024expertqa}. 

In contrast, when the potential evidence is a single long document that fits the LLM context window, the confounding retrieval can be omitted, as LLMs can cite from their input \cite{gao2023alce}. Still, it is possible that separating response generation and evidence retrieval improves performance, as shown in related works on task decomposition \cite{sun2023pearl} and reducing long LLM inputs, where \citet{agrawal2024cant} found positive effects, while \citet{xu2024retrieval} and \citet{bai2023longbench} did not. These works did not consider attribution, so there is a lack of knowledge on the effect of separating response generation and evidence retrieval, and the optimal approach to attribution on long documents. 

The document length poses additional challenges: Recent works have found that LLM performance on long-input tasks depends on the position of the information in the context \cite{liu2024lost, staniszewski2024structured, ravaut2024context}. Whether this can also be observed for attribution has not yet been investigated.

Evidence quality can be automatically evaluated without external reference \cite{honovich2022true, yue-etal-2023-automatic, tang2024minicheck}. If evidence quality were positively correlated with response quality, bad responses could be filtered by identifying responses with bad evidence quality, thereby improving abstaining. If not, this would lead to abstaining from potentially helpful responses with insufficient evidence. Current research on the relation of response quality and evidence quality is inconclusive: \citet{bohnet2022attributed} and \citet{gao2023alce} reranked multiple sampled responses by evidence quality. While the former found an improvement in response quality, the latter did not. Neither provide an analysis, so we lack understanding if and how evidence quality correlates with response quality.


To close these gaps,  we compile \benchmarkname{}, a \textbf{L}ong-document \textbf{A}ttribution \textbf{B}enchmark of 6 long-document datasets with diverse tasks (QA, classification, fact checking, NLI and summarization) and domains (science, law, governmental and Wikipedia). We conduct experiments using approaches to attribution with and without additional retrieval on 5 LLMs of varying sizes, prompted and fine-tuned, to answer three research questions: 

\textbf{RQ1: What are optimal approaches for attribution in long document tasks?} We find that large and fine-tuned models reach best evidence quality by directly citing from input documents, but small prompted LLMs can benefit from post-hoc evidence retrieval.

\textbf{RQ2: Do LLMs exhibit positional biases in evidence attribution?} Concerning evidence retrieval, except for GovReport, we find no particular bias, as the predicted and gold evidence distributions are mostly similar. However, we find that response quality generally decreases as evidence appears later in the document.

\textbf{RQ3: What is the relation between evidence quality and response quality?} We find that evidence quality can predict response quality on datasets with single-fact responses, but not so for multi-fact responses, as models struggle with providing evidence for complex claims.

\section{Related Work}
\label{sec:related_work}
\paragraph{Attribution}

Current research in attribution is done in three strains: First, some works evaluate a range of approaches in their ability to produce attributed responses \cite{bohnet2022attributed, liu-etal-2023-evaluating}, some proposing new datasets \cite{malaviya2024expertqa} or benchmarks \cite{deyoung-etal-2020-eraser, gao2023alce}. Second, methodological works propose new fine-tuning \cite{schimanski2024faithful, huang2024training} and prompting \cite{berchansky2024cotar, fierro2024learning} methods to improve the citation capabilities of language models, but do not compare to approaches with additional retrieval. Both of these strains have focused on open domain QA, and neglected the long document scenario. Here, we close these gaps by providing a comprehensive investigation of attribution for long documents. \\
To evaluate evidence quality automatically, most works have used TRUE, Flan-T5-XXL fine-tuned on several NLI datasets \cite{honovich2022true, bohnet2022attributed, gao2023alce, fierro2024learning, huang2024training}. More recently, Attrscore \cite{yue-etal-2023-automatic} and Minicheck \cite{tang2024minicheck} were proposed specifically for the evaluation of attributability. We compare these models and employ the best-performing for attributability evaluation.

\paragraph{LLMs for long documents} While there is no universal definition of "long documents", existing long document benchmarks contain documents of 1500 to 50000 words average length \cite{shaham-etal-2023-zeroscrolls, dong-etal-2024-bamboo-comprehensive, an2023leval, li2023loogle}. Initial LLMs were limited to contexts of less than 2000 tokens \cite{brown2020fewshot, touvron2023llama}, but recent advances in hardware and efficiency \cite{dao2023flashattention2} have spurred the development of models with context $\geq$$8000$ tokens, e.g. Longchat \cite{longchat2023}, Mistral \cite{jiang2023mistral}, GPT-3.5-16K \footnote{\url{https://platform.openai.com/docs/models/gpt-3-5-turbo}} or GPT-4-Turbo-128K.\footnote{\url{https://platform.openai.com/docs/models/gpt-4-turbo-and-gpt-4}} We add to this line of research by evaluating a range of models in their attribution capabilities.


\section{Formalization}
\label{sec:formalization}
We first define the task of producing attributed responses. Based on this, we define the approaches to attribution compared in this work.

\paragraph{Task definition} We assume an instruction $I$ (e.g. a question) and a document $D$  consisting of  segments $d$ (e.g. paragraphs). There are two subtasks that can be solved jointly or independently: One is to generate a \textit{response} $R$ (e.g. an answer) containing statements $r$. If the response is not abstained (e.g. by saying ``unanswerable''), the other subtask is to retrieve \textit{evidence} $E_i \subset D$ for each $r_i$ such that $r_i$, is \textit{attributable} to $E_i$, i.e. "according to $E_i$, $r_i$" is true \cite{rashkin2023attribution}.

\paragraph{Approaches to attribution} Different approaches to attribution can be defined based on the subtask order (Fig.~\ref{fig:attribution}): (1) \texttt{post-hoc}: an LLM generates a response $R$, and evidence is retrieved from $D$ based on $R$. (2) \texttt{retrieve-then-read}: Evidence $E$ is retrieved from $D$, and an LLM generates a response based on $E$. (3) \texttt{citation}: Based on $D$, an LLM generates a response and retrieves evidence in one step. \\
To decrease the input length in \texttt{post-hoc} and \texttt{citation}, $D$ can be reduced to $D' \subset D$ via an additional initial retrieval step. This results in two further attribution approaches: (4) \texttt{reduced-post-hoc} and (5) \texttt{reduced-}\texttt{citation}.

\section{Methods}
\label{sec:methods}
We first introduce the datasets (§\ref{sec:methods/datasets}) and evaluation metrics (§\ref{sec:methods/evaluation}) used in \benchmarkname{}, our long document attribution benchmark. We then describe our experimental setup for generation (§\ref{sec:methods/modeling/generation}) and retrieval (§\ref{sec:methods/modeling/retrieval}).

\subsection{Datasets}
\label{sec:methods/datasets}

\begin{table*}
    \centering
    \small
\begin{tabular}{l|lllllll}
\toprule
                        & Domain        & Task          & \#Inst Train/Dev/Test & Doc \#W & Res $|R|$/\#W & \#Evi & Evi Lvl \\ \midrule
QASPER (QSP) [1]            & Science       & QA            & 2675/1005/1451      & 3937       & 1/12       & 1.7  & para        \\
Natural Questions (NQ) [2] & Wiki          & QA            & 232191/6205/7307      & 4693       & 1/3       & 1     & para     \\
Evidence Inference (EI) [3] & Science       & Cls           & 18545/1232/1218      & 3962       & 1/1       & 1.1     & para     \\
Wice (WIC) [4]             & Web           & FC & 4234/349/358      & 1339       & 1/1       & 3.7    & sent      \\
ContractNLI (CNLI) [5]     & Legal         & NLI           & 7191/1037/2091      & 1697       & 1/1       & 1.5   & para       \\
GovReport (GR) [6]         & US Gov. & Sum           & 15107/964/969      & 8464       & 20/517       & N/A       & N/A \\
\bottomrule
\end{tabular}
    \caption{The datasets in \benchmarkname{} span multiple domains and task types. The numbers for Natural Questions differ from the original publication, as we filtered the instances (§\ref{sec:appendix/datasets}). Column names: Doc \#W: Average number of words per document. Res $|R|$/\#W: Number of statements / Average number of words per response. \#Evi: Number of annotated evidence segments per instance. Evi Lvl: Level of annotated evidence. Tasks: QA: Question answering. Cls: Classification. FC: Fact checking. NLI: Natural Language Inference. Sum: Summarization. [1] \citet{dasigi-etal-2021-dataset} [2] \citet{kwiatkowski2019natural} [3] \citet{deyoung-etal-2020-evidence} [4] \citet{kamoi-etal-2023-wice} [5] \citet{koreeda-manning-2021-contractnli-dataset} [6] \citet{huang-etal-2021-efficient}}
    \label{tab:datasets}
\end{table*}

The datasets in \benchmarkname{} are shown in Table~\ref{tab:datasets}. All datasets are in English. GovReport \cite{huang-etal-2021-efficient} is the only dataset without annotated gold evidence. To simulate gold evidence, we use BM25 to find the 2 best-matching paragraphs from a document for each sentence in the gold summary similar to \citet{ravaut2024context}. Due to limited resources, we use at most 2000 test instances from any dataset (100 for GPT-4). For details and examples see §\ref{sec:appendix/datasets}. 


\subsection{Evaluation}
\label{sec:methods/evaluation}
We evaluate two main properties of attributed responses: 1. \textit{Response quality}, i.e. whether a model-generated response is correct in reference to an annotated ground-truth. 2. \textit{Evidence quality}, i.e. whether the provided evidence is relevant to the given response. These are evaluated with task-specific metrics (see below). For QA datasets, we additionally evaluate \textit{Unanswerable F1}, i.e. whether models abstain from responding to unanswerable instances.

\subsubsection{Response Quality} 

For comparability with related work, we used established metrics to evaluate response quality: Exact match F1 (EM)\footnote{Found to correlate better than BERTScore \cite{Zhang*2020BERTScore:} with human judgment in our preliminary experiments.} on QASPER \cite{dasigi-etal-2021-dataset, shaham-etal-2023-zeroscrolls} and Natural Questions \cite{kwiatkowski2019natural, bohnet2022attributed}, classification macro F1 (CF1) for Evidence Inference \cite{deyoung-etal-2020-evidence}, Wice \cite{kamoi-etal-2023-wice} and ContractNLI \cite{koreeda-manning-2021-contractnli-dataset} and ROUGE-L\footnote{ROUGE has shown good correlation with human relevance judgments in \citet{wu-etal-2024-less}.} (RL, \citealt{lin-2004-rouge}) for GovReport \cite{huang-etal-2021-efficient, shaham-etal-2023-zeroscrolls}.

\subsubsection{Evidence Quality}

\paragraph{Evidence F1 (EF1)} For datasets that define a \textit{fixed} response vocabulary (i.e. Evidence Inference, Wice and ContractNLI), we compute evidence quality as evidence F1, comparing the predicted evidence with annotated ground truth evidence. If there is no annotated evidence, evidence F1 is 1 if no evidence was predicted and otherwise 0.

\paragraph{Attributability (ATT)} For the other datasets, evidence F1 is insufficient: GovReport does not come with annotated evidence, and for datasets with \textit{free-form} responses (QASPER and Natural Questions), evidence F1 is too rigid: A model might produce a response different from the ground truth, but supported by the retrieved evidence. Even though the predicted evidence is relevant to the predicted response, evidence F1 might be low. For these datasets, we evaluate evidence quality as attributability\footnote{Also known as citation recall \cite{gao2023alce}.} \cite{rashkin2023attribution}, \cite{gao2023alce, huang-etal-2021-efficient, schimanski2024faithful}. We assume an attributability evaluation model  $\mathrm{M_a}(E, r) \rightarrow \{1, 0\}$. The attributability score (AS) of a response is computed as the proportion of attributable statements. 
\begin{equation}
    \mathrm{ATT} = \frac{1}{n}\sum_{i=1}^n \mathrm{M_a}(E_i,r_i) \nonumber
\end{equation}
On QA tasks, models can abstain from responding. In these cases, we do not evaluate attributability, as no evidence is required. Because this could encourage abstaining from responding too often, we additionally evaluate Unanswerable F1 (see below). 

\subsubsection{Attributability Evaluation Model Selection}
\label{sec:methods/evaluation/attributability_eval}

To select a model for attributability evaluation, we created test datasets for QASPER, Natural Questions and GovReport, and evaluated TRUE \cite{honovich2022true}, AttrScore \cite{yue-etal-2023-automatic} and Minicheck \cite{tang2024minicheck}. For AttrScore, we map "Contradictory" and "Extrapolatory" predictions to a single "not attributable" class. 

\paragraph{Human annotation} We generated attributed responses using GPT 3.5-\texttt{post-hoc} with BM25 for evidence retrieval. Two authors of this study, both holders of a Master's degree and fluent in English, annotated attributability (attributable or not attributable) for 200 responses (200 sentences for GovReport) and reached an agreement of 0.74 (QASPER), 0.77 (Natural Questions) and 0.76 (Govreport) Cohen's $\kappa$. 

\paragraph{Results} We report accuracy \cite{gao2023alce} and balanced accuracy \cite{tang2024minicheck} scores in Table~\ref{tab:attr_eval}. The scores are comparable to \citet{gao2023alce}, who reported 85\% accuracy and \citet{tang2024minicheck}, who reported between 59\% and 84\% balanced accuracy on their respective benchmarks. We select the best-performing model to evaluate attributability: Minicheck for QASPER, and TRUE for Natural Questions and GovReport. 

\subsubsection{Unanswerable F1 (UF1)} We set this up as a classification task similar to \citet{slobodkin2023curious}. Gold labels are determined depending on the number of annotations \cite{kwiatkowski2019natural}: For 3 annotations or less: ``Unanswerable'' if all annotators annotated unanswerable, else ``answerable''. More than 3 annotations: ``Unanswerable'' if at most one annotator did not annotate unanswerable, else ``answerable''. If a model abstains, its prediction is set to ``unanswerable'', else ``answerable''. \\
To detect abstaining, we compiled a list of keyphrases based on \citet{slobodkin2023curious} and manual inspection. If a response contained a keyphrase, any predicted evidence was removed, and the response was set to "unanswerable" (see §\ref{sec:appendix/technical_details}).

\begin{table}[]
    \centering
    \small
    \begin{tabular}{l|ccc}
    \toprule
\textbf{Model} & \textbf{QSP} & \textbf{NQ} & \textbf{GR} \\ \midrule
TRUE           &   79/80            & \textbf{83/83}              &  \textbf{79/78}       \\
AttrScore      &       76/71          &  68/67             &  76/52  \\
Minicheck      & \textbf{83/82} & 82/82 & 79/70 \\
\bottomrule
\end{tabular}
    \caption{Attributability evaluation model selection. Metrics: accuracy / balanced accuracy \cite{tang2024minicheck}}
    \label{tab:attr_eval}
\end{table}

\subsection{Generation}
\label{sec:methods/modeling/generation}

\paragraph{Model selection} We focus on two groups of LLMs with at least 8K tokens input length: (1) The large state of the art models GPT-3.5\footnote{gpt-35-turbo-0613-16k} and GPT-4\footnote{gpt-4-turbo-128k}, as they hold top positions in other long document benchmarks \cite{li2023loogle, shaham-etal-2023-zeroscrolls, an2023leval} (2) Small (\textasciitilde{}3-7B) models that are accessible with limited resources. We tested a range of models with \texttt{citation} on QASPER and GovReport and selected Longchat\footnote{\href{https://huggingface.co/lmsys/longchat-7b-v1.5-32k}{Longchat-7B-v1.5-32K}} \cite{longchat2023} and Mistral\footnote{\href{https://huggingface.co/mistralai/Mistral-7B-Instruct-v0.2}{Mistral-7B-Instruct-v0.2}} \cite{jiang2023mistral} as the best-performing for prompting and Flan-T5 \footnote{\href{https://huggingface.co/google/flan-t5-xl}{Flan-T5-XL}} \cite{pmlr-v202-longpre23a} as the best performing for fine-tuning. For complete selection results and hyperparameters see §\ref{sec:appendix/generation_model}.

\paragraph{Prompts} We employ separate prompt sets for \texttt{citation} and \texttt{non-citation}. Similar to \citet{shaham-etal-2023-zeroscrolls}, we keep instructions short, including guidance on the expected responses and output format. Prompts contained three in-context examples, where documents were shortened to title, section headings and annotated evidence. For details and prompt optimization experiments, see §\ref{sec:appendix/prompts}. 

\paragraph{Response parsing} For all datasets except GovReport, responses consist of single statements. For GovReport, we split responses into statements (sentences) using NLTK \cite{bird-2006-nltk}. For \texttt{citation}, LLMs are expected to generate segment identifiers (``[1] [2]'') at the end of each statement.

\subsection{Retrieval}
\label{sec:methods/modeling/retrieval} 

\paragraph{Retriever selection} For retrieval in \texttt{post-hoc}, \texttt{retrieve-then-read} and \texttt{reduced}, we employ sparse and dense retrievers that showed good performance in related work\footnote{For efficiency reasons, we do not use LLMs for retrieval.} \cite{thakur2021beir}: BM25 \cite{robertson2009bm25, gao2023alce}, GTR \cite{ni2021large, gao2023alce, bohnet2022attributed}, Contriever \cite{izacard2022unsupervised, xu2024retrieval, bai2023longbench}, Dragon \cite{lin2023train, xu2024retrieval} and the best-performing Sentence Transformer ``all-mpnet-base-v2''\footnote{\url{https://sbert.net/docs/sentence_transformer/pretrained_models.html\#semantic-search-models}} \cite{reimers2019sentencebert}. For each combination of approach and task, we selected the best-performing retriever using GPT-3.5 as response generator (see §\ref{sec:appendix/retrieval}).

\paragraph{Details} In \texttt{retrieve-then-read} and \texttt{reduced}, queries were constructed based on information available in the instruction (e.g. question or a claim). For GovReport, similar to \citet{zhang2020pegasus}, we created queries from all document segments and retrieved paragraphs based on self-similarity. In \texttt{post-hoc}, queries were constructed based on the instruction and the generated response. For both \texttt{post-hoc} and \texttt{retrieve-then-read}, we retrieve 5 evidence segments for Wice, and 2 for all other datasets based on evidence statistics (see Table~\ref{tab:datasets}). In \texttt{reduced} approaches we reduce input documents to 10 segments based on \cite{xu2024retrieval} (see §\ref{sec:appendix/retrieval}).

\section{Experiments}
\label{sec:experiments}

\begin{table*}
\centering 
\begin{adjustbox}{width=\linewidth,center}
\begin{tabular}{ll|ccc@{\hskip 0.25in}ccc@{\hskip 0.25in}cc@{\hskip 0.25in}cc@{\hskip 0.25in}cc@{\hskip 0.25in}cc|cc}
\toprule
 &  & \multicolumn{3}{c@{\hskip 0.25in}}{QSP} & \multicolumn{3}{c@{\hskip 0.25in}}{NQ} & \multicolumn{2}{c@{\hskip 0.25in}}{EI} & \multicolumn{2}{c@{\hskip 0.25in}}{WIC} & \multicolumn{2}{c@{\hskip 0.25in}}{CNLI} & \multicolumn{2}{c@{\hskip 0.25in}}{GR} & \multicolumn{2}{c@{\hskip 0.25in}}{Avg} \\
 &  & \cellcolor{blue}EM & \cellcolor{green}ATT & UF1 & \cellcolor{blue}EM & \cellcolor{green}ATT & UF1 & \cellcolor{blue}CF1 & \cellcolor{green}EF1 & \cellcolor{blue}CF1 & \cellcolor{green}EF1 & \cellcolor{blue}CF1 & \cellcolor{green}EF1 & \cellcolor{blue}RL & \cellcolor{green}ATT & \cellcolor{blue}RQ & \cellcolor{green}EQ \\
\midrule
\multirow[t]{5}{*}{\textbf{GPT-3.5}} & \textbf{\texttt{p-h}} & 45 & 62 & 65 & \textbf{51} & 44 & \textbf{58} & 77 & 22 & \textbf{52} & 48 & 44 & 40 & \textbf{26} & \textbf{74} & 49.17 & 48.33 \\
\textbf{} & \textbf{\texttt{rtr}} & 42 & \textbf{78} & 51 & 50 & 42 & 57 & 73 & 25 & 50 & 44 & 47 & 41 & 21 & 40 & 47.17 & 45.00 \\
\textbf{} & \textbf{\texttt{cit}} & \textbf{51} & 55 & \textbf{70} & 47 & 38 & 53 & \textbf{78} & 50 & \textbf{52} & 59 & 43 & 53 & \textbf{26} & 59 & \textbf{49.50} & 52.33 \\
\textbf{} & \textbf{\texttt{r-p-h}} & 48 & 71 & 58 & 50 & \textbf{45} & 57 & \textbf{78} & 22 & \textbf{52} & 48 & 45 & 40 & 22 & 64 & 49.17 & 48.33 \\
\textbf{} & \textbf{\texttt{r-cit}} & 50 & 69 & 64 & 46 & 38 & 53 & \textbf{78} & \textbf{52} & 48 & \textbf{63} & \textbf{48} & \textbf{57} & 22 & 58 & 48.67 & \textbf{56.17} \\
\cline{1-18}
\multirow[t]{5}{*}{\textbf{GPT-4}} & \textbf{\texttt{p-h}} & 65 & 68 & 68 & 52 & 42 & 56 & \textbf{87} & 24 & 32 & 36 & \textbf{66} & 50 & \textbf{27} & \textbf{63} & 54.83 & 47.17 \\
\textbf{} & \textbf{\texttt{rtr}} & 56 & \textbf{83} & 56 & \textbf{58} & 36 & \textbf{61} & 74 & 31 & 22 & 28 & 64 & 59 & 20 & 27 & 49.00 & 44.00 \\
\textbf{} & \textbf{\texttt{cit}} & \textbf{68} & 76 & \textbf{71} & 51 & \textbf{49} & 57 & 86 & \textbf{64} & \textbf{35} & \textbf{47} & 63 & 64 & \textbf{27} & 62 & \textbf{55.00} & \textbf{60.33} \\
\textbf{} & \textbf{\texttt{r-p-h}} & 62 & 71 & 69 & 51 & 43 & 57 & 84 & 24 & 28 & 30 & 65 & 50 & 22 & 54 & 52.00 & 45.33 \\
\textbf{} & \textbf{\texttt{r-cit}} & 63 & 76 & 66 & 54 & \textbf{49} & 58 & 83 & 49 & 33 & 46 & 64 & \textbf{67} & 22 & 55 & 53.17 & 57.00 \\
\cline{1-18}
\multirow[t]{5}{*}{\textbf{Flan-T5}} & \textbf{\texttt{p-h}} & \textbf{55} & 67 & \textbf{58} & 81 & 61 & 74 & 86 & 22 & 44 & 46 & \textbf{77} & 57 & \textbf{27} & \textbf{90} & 61.67 & 57.17 \\
\textbf{} & \textbf{\texttt{rtr}} & 43 & \textbf{74} & 55 & 79 & 62 & 74 & 77 & 25 & 43 & 43 & 64 & 58 & 19 & 67 & 54.17 & 54.83 \\
\textbf{} & \textbf{\texttt{cit}} & 53 & 57 & 56 & \textbf{84} & \textbf{75} & 71 & 83 & \textbf{59} & \textbf{53} & \textbf{70} & \textbf{77} & \textbf{78} & 25 & 71 & \textbf{62.50} & \textbf{68.33} \\
\textbf{} & \textbf{\texttt{r-p-h}} & 53 & 72 & \textbf{58} & 80 & 62 & \textbf{75} & \textbf{87} & 22 & 45 & 46 & 76 & 58 & 23 & 84 & 60.67 & 57.33 \\
\textbf{} & \textbf{\texttt{r-cit}} & 52 & 62 & 55 & \textbf{84} & \textbf{75} & 73 & 84 & 57 & \textbf{53} & 68 & 75 & 75 & 22 & 68 & 61.67 & 67.50 \\
\cline{1-18}
\multirow[t]{5}{*}{\textbf{Longchat}} & \textbf{\texttt{p-h}} & 23 & 57 & \textbf{62} & 21 & 33 & 38 & \textbf{66} & 22 & 23 & \textbf{31} & 38 & 33 & 24 & \textbf{67} & \textbf{32.50} & \textbf{40.50} \\
\textbf{} & \textbf{\texttt{rtr}} & 29 & 56 & 49 & \textbf{27} & 32 & \textbf{43} & 43 & \textbf{25} & 19 & 25 & \textbf{39} & \textbf{39} & 21 & 41 & 29.67 & 36.33 \\
\textbf{} & \textbf{\texttt{cit}} & 21 & 6 & 60 & 17 & 2 & 38 & \textbf{66} & 13 & \textbf{24} & 19 & 36 & 14 & \textbf{25} & 9 & 31.50 & 10.50 \\
\textbf{} & \textbf{\texttt{r-p-h}} & \textbf{33} & \textbf{58} & 54 & 25 & \textbf{35} & 41 & 61 & 22 & 20 & 27 & 33 & 30 & 22 & 63 & 32.33 & 39.17 \\
\textbf{} & \textbf{\texttt{r-cit}} & 26 & 26 & 53 & 21 & 4 & 42 & 65 & 13 & 20 & 25 & 24 & 12 & 22 & 23 & 29.67 & 17.17 \\
\cline{1-18}
\multirow[t]{5}{*}{\textbf{Mistral}} & \textbf{\texttt{p-h}} & 32 & 67 & 59 & 28 & 42 & 43 & 75 & 20 & 37 & 45 & 41 & 40 & \textbf{22} & 63 & 39.17 & 46.17 \\
\textbf{} & \textbf{\texttt{rtr}} & 27 & \textbf{76} & 50 & \textbf{30} & 42 & \textbf{46} & 66 & 25 & 37 & 42 & \textbf{51} & 43 & 20 & 39 & 38.50 & 44.50 \\
\textbf{} & \textbf{\texttt{cit}} & 35 & 39 & \textbf{61} & 26 & 15 & 41 & 75 & \textbf{27} & 34 & 39 & 47 & 38 & \textbf{22} & 0 & 39.83 & 26.33 \\
\textbf{} & \textbf{\texttt{r-p-h}} & 31 & 71 & 58 & 29 & \textbf{44} & 43 & 76 & 21 & \textbf{38} & 45 & 45 & 38 & 21 & \textbf{64} & 40.00 & \textbf{47.17} \\
\textbf{} & \textbf{\texttt{r-cit}} & \textbf{36} & 53 & 58 & 28 & 21 & 42 & \textbf{77} & 26 & 35 & \textbf{47} & \textbf{51} & \textbf{44} & \textbf{22} & 1 & \textbf{41.50} & 32.00 \\
\cline{1-18}
\bottomrule
\end{tabular}
\end{adjustbox}
\caption{Evaluation on \benchmarkname{}, all scores show percentages. \texttt{Citation} / \texttt{reduced-citation} mostly perform best, with notable exceptions for GovReport and Longchat (see §\ref{sec:experiments/optimal_methods_for_attribution}). \texttt{p-h}: \texttt{post-hoc}. \texttt{rtr}: \texttt{retrieve-then-read}. \texttt{cit}: \texttt{citation}. \texttt{r-}: \texttt{reduced-}. EM: Exact match F1, ATT: Attributability, UF1: Unanswerable F1, CF1: Classification F1, EF1: Evidence F1, RL: Rouge-L, RQ (EQ): Average response (evidence) quality, mean of all scores with \colorbox{blue}{blue} (\colorbox{green}{green}) shade.}
\label{tab:main_table}

\end{table*}


\subsection{RQ1: What are optimal approaches to attribution in long document tasks?}
\label{sec:experiments/optimal_methods_for_attribution}

Table~\ref{tab:main_table} shows the results from all combinations of selected models. Due to the large number of experiments, all results are from single runs. 

\paragraph{Which approach produces the highest evidence quality?} Flan-T5-XL has higher average scores than GPT-3.5 and GPT-4, while the Longchat and Mistral scores are lower. For GPT-3.5, GPT-4 and Flan-T5, \texttt{citation} / \texttt{reduced-citation} results in the best evidence quality on average and most datasets, and \texttt{retrieve-then-read} performs worst. \texttt{Post-hoc} works best for Longchat and Mistral, and for all models on GovReport.

\paragraph{Does \texttt{citation} hurt response quality?} It could be assumed that \texttt{post-hoc} results in better response quality than \texttt{citation}, as task decomposition can improve performance \cite{gao-etal-2023-rarr}. We compare average response quality between (\texttt{reduced-})\texttt{citation} and (\texttt{reduced-})\texttt{post-hoc} for GPT-3.5, GPT-4 and Flan-T5-XL. In no case, the response quality for \texttt{citation} is more than 0.5 points lower than for \texttt{post-hoc}, showing that \texttt{citation} has a minimal effect on response quality.

\paragraph{Does performance depend on document length?} Tables \ref{tab:response_quality_vs_doc_length_correlation} and \ref{tab:evidence_quality_vs_doc_length_correlation} show the correlation between response quality or evidence quality and document length, which is mostly negative. A notable exception is GPT-4 evaluated for response quality, where correlation with document length is positive on all tasks but GovReport.

\paragraph{Does reduction of the input document help?} Comparing \texttt{reduced-post-hoc} to \texttt{post-hoc} and \texttt{reduced-citation} to \texttt{citation}, we find that response quality is mostly better for the non-\texttt{reduced} variant, with the exception of Mistral. Regarding evidence quality, reduction only helps for GPT-3.5-\texttt{citation}, Longchat-\texttt{citation} and -\texttt{post-hoc} and Mistral-\texttt{citation}.

\paragraph{Discussion} \texttt{Citation} or \texttt{reduced-citation} result in the best average evidence quality, while not hurting response quality, in line with recent work showing LLM capabilities for retrieval \cite{ma2023zeroshot}. The GovReport task and the small models Longchat and Mistral are exceptions to this, as \texttt{post-hoc} results in better evidence quality in these cases. For GovReport, the higher evidence quality with \texttt{post-hoc} can be explained with the "repetitive" nature of the summarization task, since the high overlap between response statements and document provides good conditions for retrievers to find evidence. For the small models, related work has shown that they lack instruction following capability to perform evidence extraction \cite{gao2023alce, schimanski2024faithful}, making \texttt{post-hoc} the better approach for the model. \\
Comparing models, fine-tuned Flan-T5-XL has higher average scores for response and evidence quality than the large prompted models GPT-3.5 and GPT-4. This could also be observed in related work \cite{huang2024training, schimanski2024faithful}. \\
Regarding the relation between input length and performance, we mostly found negative correlation, similar to \citet{bai2023longbench} and \citet{kwan2024m4le}.
Still, similar to \citet{xu2024retrieval} and \citet{bai2023longbench}, we have not found a general beneficial effect of input reduction on performance. The positive results from \citet{agrawal2024cant} were obtained in a multi-document setting, where there is no logical coherence between the input documents. In contrast, the logical coherence in long documents can be disrupted through reduction, which can make processing of the reduced document more difficult.  

\subsection{RQ2: Do LLMs exhibit positional biases in attribution?}
\label{sec:experiments/positional_biases}

\begin{figure*}
    \centering
    \includegraphics{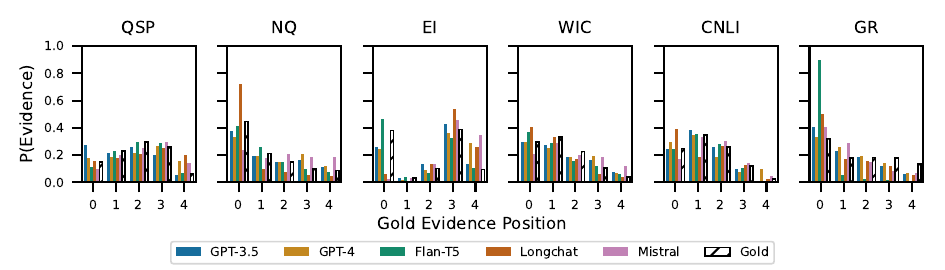}
    \label{fig:evidence_position_gold_vs_predicted}
    \centering
    \includegraphics{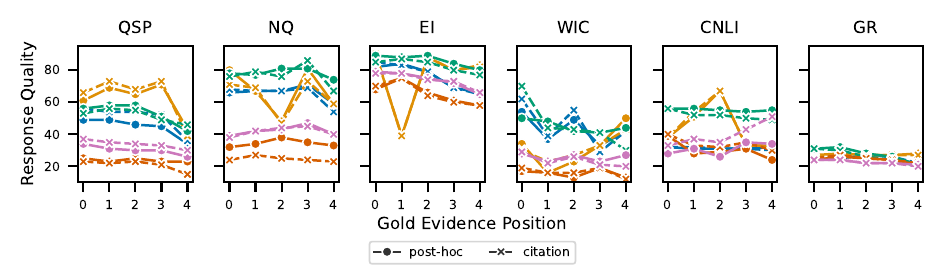}
    \caption{Top: Evidence distribution (predicted via \texttt{citation}) by position in the document. Except for GovReport, no positional bias is visible. Bottom: Response quality by position of gold evidence in the document. Negative correlation between evidence position and response quality is visible in several cases (Table~\ref{tab:evidence_position_performance}) (see §\ref{sec:experiments/positional_biases}).}
    \label{fig:evidence_position}
\end{figure*}


Several works have shown that LLM performance depends on the position of information in the input \cite{liu2024lost, staniszewski2024structured, ravaut2024context}. We investigate whether this phenomenon exists for attribution.

\paragraph{Do predicted and gold evidence distributions agree?} Figure~\ref{fig:evidence_position} (top) shows the predicted evidence distributions from \texttt{citation}\footnote{We focus on \texttt{citation} because only in this approach the LLM performs evidence retrieval.} and the gold distributions.\footnote{For GovReport, we matched summary sentences to paragraphs using BM25 to obtain gold evidence, see §\ref{sec:methods/datasets}} The models generally follow the gold evidence distribution, with Longchat showing the strongest deviation. GovReport is an exception: all models show a higher focus on the beginning of the document, especially Flan-T5-XL. Figures \ref{fig:gold_evidence_position_evidence_quality} and \ref{fig:predicted_evidence_position_evidence_quality} show the relation of evidence quality and the position of gold or predicted evidence, respectively. Neither of them show a clear dependence of evidence quality on evidence position.

\paragraph{Does response quality depend on the position of gold evidence?} We grouped instances by evidence position and evaluated the approaches with full document input (\texttt{citation} and \texttt{post-hoc}) separately for each group (Fig.~\ref{fig:evidence_position}, bottom). The strong fluctuations of GPT-4 can be explained by the fact that it was evaluated on 100 samples only. It seems that response quality decreases as evidence appears later in the document. Similar to \citet{ravaut2024context}, we computed Spearman correlation between response quality and evidence position, and find that correlation is mostly negative, but not always significant (Table~\ref{tab:evidence_position_performance}).

\paragraph{Discussion} Except for GovReport, we could not find positional biases in LLM evidence retrieval. From the results of \citet{liu2024lost}, we expected to find a ``Lost in the Middle'' effect, i.e. a reduction of retrieved evidence or performance in the middle of documents. Rather, we found a decrease in response quality towards the end of the  document similar to \citet{ravaut2024context}. We and \citet{ravaut2024context} work with coherent long documents, while \citet{liu2024lost} work in a RAG-like setup, where the input are multiple documents without coherence, and models might expect an ordering by relevance. This might explain the different results.




\subsection{RQ3: What is the relation between evidence quality and response quality?}
\label{sec:experiments/evidence_response}


\begin{table*}[]
    \centering
    \small
    \begin{tabular}{ll||cc|cc|c|c|c|c}
\toprule
 &  & \multicolumn{2}{c|}{QSP} & \multicolumn{2}{c|}{NQ} & EI & WIC & CNLI & GR \\
 &  & EM & UF1 & EM & UF1 & CF1 & CF1 & CF1 & RL \\
\midrule
\multirow[t]{2}{*}{\textbf{GPT-3.5}} & \textbf{post-hoc} & 3 & 2 & 17 & 20 & 12 & 4 & 19 & 0 \\
\textbf{} & \textbf{citation} & 2 & 1 & 14 & 18 & 13 & 4 & 25 & -1 \\
\cline{1-10}
\multirow[t]{2}{*}{\textbf{GPT-4}} & \textbf{post-hoc} & 4 & 1 & 20 & 27 & 7 & 10 & 16 & 0 \\
\textbf{} & \textbf{citation} & 4 & 3 & 6 & 18 & 10 & 7 & 20 & 1 \\
\cline{1-10}
\multirow[t]{2}{*}{\textbf{Flan-T5}} & \textbf{post-hoc} & 0 & 2 & 13 & 12 & 7 & 1 & 8 & 0 \\
\textbf{} & \textbf{citation} & 2 & 1 & 11 & 10 & 11 & 3 & 8 & -1 \\
\cline{1-10}
\multirow[t]{2}{*}{\textbf{Longchat}} & \textbf{post-hoc} & 6 & 2 & 13 & 20 & 18 & 16 & 26 & 0 \\
\textbf{} & \textbf{citation} & 4 & 1 & 4 & 7 & 10 & 11 & 15 & 1 \\
\cline{1-10}
\multirow[t]{2}{*}{\textbf{Mistral}} & \textbf{post-hoc} & 6 & 2 & 13 & 19 & 9 & 14 & 23 & 1 \\
\textbf{} & \textbf{citation} & 6 & 1 & 9 & 12 & 9 & 12 & 15 & 0 \\
\bottomrule
\end{tabular}
    \caption{Difference in response quality-coverage AUC between responses ordered by evidence quality (attributability) and random ordering. Evidence quality predicts response quality on several datasets (see §\ref{sec:experiments/evidence_response})}
    \label{tab:selective_prediction}
\end{table*}
\begin{figure*}
    \includegraphics[trim={0 .5cm 0 .3cm},clip]{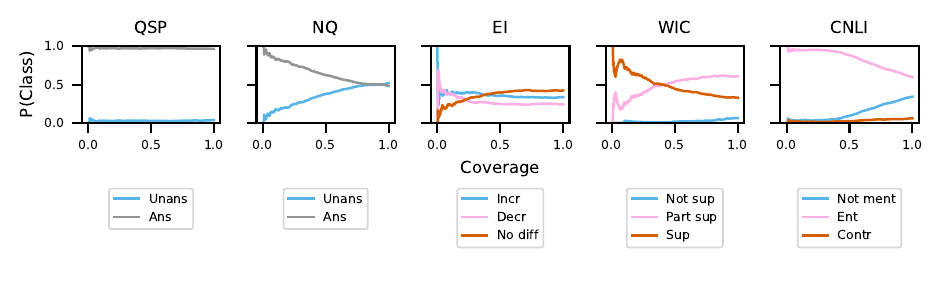}
    \caption{Gold class distribution in selective prediction for GPT-3.5-\texttt{citation}.}
    \label{fig:selective_prediction}
\end{figure*}




Attributability evaluation models \cite{honovich2022true, yue-etal-2023-automatic, tang2024minicheck} can evaluate evidence quality without external reference. If evidence quality were positively correlated with response quality, this could be used to abstain from low-quality responses. We test this with \textit{selective prediction} \cite{yaniv2010selective}, using evidence quality scores from attributability evaluation as an estimate of \textit{confidence}.

\paragraph{Selective Prediction} We begin by filtering predicted responses that don't require evidence\footnote{``Unanswerable'' on QASPER and Natural Questions, ``not supported'' on Wice, ``not mentioned'' on ContractNLI.}, as attributability cannot be evaluated on these. We sort the remaining $l$ predictions by evidence quality\footnote{We evaluated evidence quality as attributability for all datasets, as it does not require annotated evidence. We used Minicheck\cite{tang2024minicheck} for QASPER and TRUE \cite{honovich2022true} for all other datasets (§\ref{sec:methods/evaluation/attributability_eval})} in descending order, obtaining the ordered set of responses $\mathcal{R}_{sel}$. For each $t\in\{0...l\}$, we compute average response quality\footnote{As filtering responses that do not require evidence produces a strong class imbalance, we use micro F1 instead of macro F1 for Wice and ContractNLI} on the subset of $\mathcal{R}_{sel}$ up to $t$ and the coverage (the proportion of responses evaluated). We evaluate the estimation of confidence through evidence quality by computing the area under the response quality-coverage curve (AUC) \cite{chen2023adaptation}. 

\paragraph{Is evidence quality a good estimate of confidence in selective prediction?}  Table \ref{tab:selective_prediction} shows the AUC difference between ordering predictions by attributability and a random order baseline (mean of 10 repeats). We see that attributability is an effective estimate of confidence on Natural Questions, Evidence Inference and ContractNLI, and to some extent on Wice. On QASPER, and GovReport, the difference to the random baseline is small. 


\paragraph{Do unanswerable instances have lower evidence quality?} For instances annotated as ``unanswerable'', ``not supported'' or ``not mentioned'', there is no annotated evidence. If models give different responses, the evidence quality should be low, and these should be filtered in selective prediction. Fig.~\ref{fig:selective_prediction} shows the distribution of gold responses in selective prediction. As expected, the proportion of such instances without sufficient evidence decreases with lower coverage for Natural Questions, Wice and ContractNLI but, surprisingly, not for QASPER.

\paragraph{Why does evidence quality fail to predict response quality?} We consider GovReport a special case, as its long responses are evaluated in their entirety,  which might be too coarse-grained to reflect the per-statement evaluation of evidence quality. This is corroborated by the fact that system-level correlation between evidence quality and response quality is significantly positive for all datasets except GovReport (Table \ref{tab:correlation}). \\
For Wice and QASPER, the possible causes are: (1) Responses are correct, but the evidence is insufficient. (2) Responses are incorrect, but the evidence is sufficient. (3) The attributability scores are wrong. We performed a manual analysis on the 50 responses with the lowest attributability from GPT-3.5-\texttt{citation}: For QASPER, the responses were often correct (answer F1 of 66), but the evidence was insufficient in 46 cases. For Wice, this could be observed in 39 cases. This implies that the LLM's failure to extract sufficient evidence is the main reason for low correlation between evidence and response quality. \\
For Wice, the attribution evaluation model failed to recognize evidence for ``partially supported'' claims in 8 cases, as it is only trained to distinguish ``supported'' and ``not supported''. This can be seen in Fig.~\ref{fig:selective_prediction}, where the proportion of ``partially supported'' decreases with lower coverage. 

\paragraph{Discussion} We explain the differences in the relationship between evidence quality and response quality by the varying dataset complexity. While Natural Questions, ContractNLI and Evidence Inference focus on single facts (e.g. a single entity or a specific contractual obligation), Wice and QASPER instances often contain multiple facts (e.g. an enumeration or  multiple subclaims), which is also reflected in the higher number of evidence segments per instance (Table~\ref{tab:datasets}). Models respond correctly, but fail to point to all necessary evidence. This is in line with related work on attribution: While \citet{bohnet2022attributed} found that response quality can be improved by attributability-based reranking on Natural Questions, \citet{gao2023alce} did not find this on their more complex benchmark.  


\section{Conclusion}
\label{sec:conclusion}
In our experiments on \benchmarkname{}, we found that \texttt{citation} is a promising approach to attribution for large or fine-tuned models, while for small prompted models, \texttt{post-hoc} extraction can improve performance. We did not find a ``Lost in the Middle'' effect, but negative correlation between evidence position and response quality in some cases. Finally, we showed that evidence quality can predict response quality for responses with low complexity. We hope that our results, code and data spur further research on long document assistants, most prominently: (1) Improving the \texttt{citation} capabilities of LLMs for complex responses. (2) Combining attributability evaluation models and iterative self-refinement \cite{gao-etal-2023-rarr} to try to improve abstained responses. 

\section{Limitations}
\paragraph{Using LLMs for retrieval} LLMs have shown good performance in reranking tasks (e.g. \citealt{ma2023zeroshot}). For efficiency reasons, we did not employ LLMs for retrieval. Instead, we employed state-of-the-art retrievers and implemented a rigorous selection procedure, elucidating the best retriever for each combination of task and approach. In the case of \texttt{reduced} approaches, we deem it unlikely to see large beneficial effects: The ``pressure'' on the retrievers was already low, as they only had to retrieve 1-3 relevant segments among 10 retrieved in total. \\
For \texttt{post-hoc} and \texttt{retrieve-then-read}, it could be that using LLMs for evidence retrieval improves performance. However this would not change our main claims: Our experiments with the \texttt{citation} approach already show that using LLMs for evidence retrieval works best. Therefore, we found the trade-off between increased computational cost and additional insights not favorable towards employing LLMs for evidence retrieval in \texttt{post-hoc} and \texttt{retrieve-then-read}. This might be different for researchers or practicioners interested in maximal performance.

\paragraph{Evaluation of evidence quality} While our attributability model selection experiments showed that these models obtain good accuracy, our analysis in \ref{sec:experiments/evidence_response} showed that edge cases are not yet handled well. Research into solving such edge cases is a promising direction for future work.

\paragraph{Datasets} We compiled a benchmark with diverse tasks, domains, response and document lengths, but naturally, we were not able to cover all variations of these properties. Many long document datasets and benchmarks are available (e.g. \citealt{li2023loogle, an2023leval, shaham-etal-2023-zeroscrolls}), but only few contain annotated evidence, which we required for the positional bias analysis in our paper (not finding a long document summarization dataset with annotated evidence, we resorted to GovReport). Given that we found only limited positional biases, extending our work to more datasets with and without annotated evidence is an interesting direction for future work. 
\label{sec:limitations}

\section{Potential Risks}
\label{sec:potential_risks}
The goal of this work is to promote the development of more trustworthy long document assistants, which we deem a promising, but low-risk research goal. 
Similarly, we deem our research methodology to be of low risk: All datasets used were created for NLP research, do not contain personal, harmful or sensitive data and were published under permissive licenses (Table~\ref{tab:dataset_licenses}). We did not employ human annotators other than the authors of the study themselves. 

\section{Acknowledgements}
\label{sec:acknowledgements}
This work was funded by the European Union (ERC, InterText, 101054961). Views and opinions expressed are however those of the authors only and do not necessarily reflect those of the European Union or the European Research Council. Neither the European Union nor the granting authority can be held responsible for them. This work has been funded by the LOEWE Distinguished Chair “Ubiquitous Knowledge Processing”, LOEWE initiative, Hesse, Germany (Grant Number: LOEWE/4a//519/05/00.002(0002)/81). We gratefully acknowledge the support of Microsoft with a grant for access to OpenAI GPT models via the Azure cloud (Accelerate Foundation Model Academic Research). 

\bibliography{main}

\begin{thebibliography}{60}
\expandafter\ifx\csname natexlab\endcsname\relax\def\natexlab#1{#1}\fi

\bibitem[{Agrawal et~al.(2024)Agrawal, Gao, and Gajek}]{agrawal2024cant}
Devanshu Agrawal, Shang Gao, and Martin Gajek. 2024.
\newblock \href {https://arxiv.org/abs/2403.05004} {Can't remember details in
  long documents? you need some r\&r}.
\newblock \emph{ArXiv preprint}, abs/2403.05004.

\bibitem[{An et~al.(2024)An, Gong, Zhong, Zhao, Li, Zhang, Kong, and
  Qiu}]{an2023leval}
Chenxin An, Shansan Gong, Ming Zhong, Xingjian Zhao, Mukai Li, Jun Zhang,
  Lingpeng Kong, and Xipeng Qiu. 2024.
\newblock \href {https://doi.org/10.18653/v1/2024.acl-long.776} {{L}-eval:
  Instituting standardized evaluation for long context language models}.
\newblock In \emph{Proceedings of the 62nd Annual Meeting of the Association
  for Computational Linguistics (Volume 1: Long Papers)}, pages 14388--14411,
  Bangkok, Thailand. Association for Computational Linguistics.

\bibitem[{Bai et~al.(2023)Bai, Lv, Zhang, Lyu, Tang, Huang, Du, Liu, Zeng, Hou,
  Dong, Tang, and Li}]{bai2023longbench}
Yushi Bai, Xin Lv, Jiajie Zhang, Hongchang Lyu, Jiankai Tang, Zhidian Huang,
  Zhengxiao Du, Xiao Liu, Aohan Zeng, Lei Hou, Yuxiao Dong, Jie Tang, and
  Juanzi Li. 2023.
\newblock \href {https://arxiv.org/abs/2308.14508} {Longbench: A bilingual,
  multitask benchmark for long context understanding}.
\newblock \emph{ArXiv preprint}, abs/2308.14508.

\bibitem[{Berchansky et~al.(2024)Berchansky, Fleischer, Wasserblat, and
  Izsak}]{berchansky2024cotar}
Moshe Berchansky, Daniel Fleischer, Moshe Wasserblat, and Peter Izsak. 2024.
\newblock \href {https://arxiv.org/abs/2404.10513} {Cotar: Chain-of-thought
  attribution reasoning with multi-level granularity}.
\newblock \emph{ArXiv preprint}, abs/2404.10513.

\bibitem[{Bohnet et~al.(2022)Bohnet, Tran, Verga, Aharoni, Andor, Soares,
  Ciaramita, Eisenstein, Ganchev, Herzig et~al.}]{bohnet2022attributed}
Bernd Bohnet, Vinh~Q Tran, Pat Verga, Roee Aharoni, Daniel Andor, Livio~Baldini
  Soares, Massimiliano Ciaramita, Jacob Eisenstein, Kuzman Ganchev, Jonathan
  Herzig, et~al. 2022.
\newblock \href {https://arxiv.org/abs/2212.08037} {Attributed question
  answering: Evaluation and modeling for attributed large language models}.
\newblock \emph{ArXiv preprint}, abs/2212.08037.

\bibitem[{Brown et~al.(2020)Brown, Mann, Ryder, Subbiah, Kaplan, Dhariwal,
  Neelakantan, Shyam, Sastry, Askell, Agarwal, Herbert{-}Voss, Krueger,
  Henighan, Child, Ramesh, Ziegler, Wu, Winter, Hesse, Chen, Sigler, Litwin,
  Gray, Chess, Clark, Berner, McCandlish, Radford, Sutskever, and
  Amodei}]{brown2020fewshot}
Tom~B. Brown, Benjamin Mann, Nick Ryder, Melanie Subbiah, Jared Kaplan,
  Prafulla Dhariwal, Arvind Neelakantan, Pranav Shyam, Girish Sastry, Amanda
  Askell, Sandhini Agarwal, Ariel Herbert{-}Voss, Gretchen Krueger, Tom
  Henighan, Rewon Child, Aditya Ramesh, Daniel~M. Ziegler, Jeffrey Wu, Clemens
  Winter, Christopher Hesse, Mark Chen, Eric Sigler, Mateusz Litwin, Scott
  Gray, Benjamin Chess, Jack Clark, Christopher Berner, Sam McCandlish, Alec
  Radford, Ilya Sutskever, and Dario Amodei. 2020.
\newblock \href
  {https://proceedings.neurips.cc/paper/2020/hash/1457c0d6bfcb4967418bfb8ac142f64a-Abstract.html}
  {Language models are few-shot learners}.
\newblock In \emph{Advances in Neural Information Processing Systems 33: Annual
  Conference on Neural Information Processing Systems 2020, NeurIPS 2020,
  December 6-12, 2020, virtual}.

\bibitem[{Chen et~al.(2023)Chen, Yoon, Ebrahimi, Arik, Pfister, and
  Jha}]{chen2023adaptation}
Jiefeng Chen, Jinsung Yoon, Sayna Ebrahimi, Sercan Arik, Tomas Pfister, and
  Somesh Jha. 2023.
\newblock \href {https://doi.org/10.18653/v1/2023.findings-emnlp.345}
  {Adaptation with self-evaluation to improve selective prediction in {LLM}s}.
\newblock In \emph{Findings of the Association for Computational Linguistics:
  EMNLP 2023}, pages 5190--5213, Singapore. Association for Computational
  Linguistics.

\bibitem[{Chiang et~al.(2023)Chiang, Li, Lin, Sheng, Wu, Zhang, Zheng, Zhuang,
  Zhuang, Gonzalez, Stoica, and Xing}]{vicuna2023}
Wei-Lin Chiang, Zhuohan Li, Zi~Lin, Ying Sheng, Zhanghao Wu, Hao Zhang, Lianmin
  Zheng, Siyuan Zhuang, Yonghao Zhuang, Joseph~E. Gonzalez, Ion Stoica, and
  Eric~P. Xing. 2023.
\newblock \href {https://lmsys.org/blog/2023-03-30-vicuna/} {Vicuna: An
  open-source chatbot impressing gpt-4 with 90\%* chatgpt quality}.
\newblock \emph{lmsys.org/blog}.

\bibitem[{Dao(2023)}]{dao2023flashattention2}
Tri Dao. 2023.
\newblock \href {https://arxiv.org/abs/2307.08691} {Flashattention-2: Faster
  attention with better parallelism and work partitioning}.
\newblock \emph{ArXiv preprint}, abs/2307.08691.

\bibitem[{Dasigi et~al.(2021)Dasigi, Lo, Beltagy, Cohan, Smith, and
  Gardner}]{dasigi-etal-2021-dataset}
Pradeep Dasigi, Kyle Lo, Iz~Beltagy, Arman Cohan, Noah~A. Smith, and Matt
  Gardner. 2021.
\newblock \href {https://doi.org/10.18653/v1/2021.naacl-main.365} {A dataset of
  information-seeking questions and answers anchored in research papers}.
\newblock In \emph{Proceedings of the 2021 Conference of the North American
  Chapter of the Association for Computational Linguistics: Human Language
  Technologies}, pages 4599--4610, Online. Association for Computational
  Linguistics.

\bibitem[{DeYoung et~al.(2020{\natexlab{a}})DeYoung, Jain, Rajani, Lehman,
  Xiong, Socher, and Wallace}]{deyoung-etal-2020-eraser}
Jay DeYoung, Sarthak Jain, Nazneen~Fatema Rajani, Eric Lehman, Caiming Xiong,
  Richard Socher, and Byron~C. Wallace. 2020{\natexlab{a}}.
\newblock \href {https://doi.org/10.18653/v1/2020.acl-main.408} {{ERASER}: {A}
  benchmark to evaluate rationalized {NLP} models}.
\newblock In \emph{Proceedings of the 58th Annual Meeting of the Association
  for Computational Linguistics}, pages 4443--4458, Online. Association for
  Computational Linguistics.

\bibitem[{DeYoung et~al.(2020{\natexlab{b}})DeYoung, Lehman, Nye, Marshall, and
  Wallace}]{deyoung-etal-2020-evidence}
Jay DeYoung, Eric Lehman, Benjamin Nye, Iain Marshall, and Byron~C. Wallace.
  2020{\natexlab{b}}.
\newblock \href {https://doi.org/10.18653/v1/2020.bionlp-1.13} {Evidence
  inference 2.0: More data, better models}.
\newblock In \emph{Proceedings of the 19th SIGBioMed Workshop on Biomedical
  Language Processing}, pages 123--132, Online. Association for Computational
  Linguistics.

\bibitem[{Dong et~al.(2024)Dong, Tang, Li, Zhao, and
  Wen}]{dong-etal-2024-bamboo-comprehensive}
Zican Dong, Tianyi Tang, Junyi Li, Wayne~Xin Zhao, and Ji-Rong Wen. 2024.
\newblock \href {https://aclanthology.org/2024.lrec-main.188} {{BAMBOO}: A
  comprehensive benchmark for evaluating long text modeling capacities of large
  language models}.
\newblock In \emph{Proceedings of the 2024 Joint International Conference on
  Computational Linguistics, Language Resources and Evaluation (LREC-COLING
  2024)}, pages 2086--2099, Torino, Italia. ELRA and ICCL.

\bibitem[{El-Yaniv and Wiener(2010)}]{yaniv2010selective}
Ran El-Yaniv and Yair Wiener. 2010.
\newblock \href {http://jmlr.org/papers/v11/el-yaniv10a.html} {On the
  foundations of noise-free selective classification}.
\newblock \emph{Journal of Machine Learning Research}, 11(53):1605--1641.

\bibitem[{Fierro et~al.(2024)Fierro, Amplayo, Huot, Cao, Maynez, Narayan, and
  Lapata}]{fierro2024learning}
Constanza Fierro, Reinald~Kim Amplayo, Fantine Huot, Nicola~De Cao, Joshua
  Maynez, Shashi Narayan, and Mirella Lapata. 2024.
\newblock \href {https://arxiv.org/abs/2404.03381} {Learning to plan and
  generate text with citations}.
\newblock \emph{ArXiv preprint}, abs/2404.03381.

\bibitem[{Gao et~al.(2023{\natexlab{a}})Gao, Dai, Pasupat, Chen, Chaganty, Fan,
  Zhao, Lao, Lee, Juan, and Guu}]{gao-etal-2023-rarr}
Luyu Gao, Zhuyun Dai, Panupong Pasupat, Anthony Chen, Arun~Tejasvi Chaganty,
  Yicheng Fan, Vincent Zhao, Ni~Lao, Hongrae Lee, Da-Cheng Juan, and Kelvin
  Guu. 2023{\natexlab{a}}.
\newblock \href {https://doi.org/10.18653/v1/2023.acl-long.910} {{RARR}:
  Researching and revising what language models say, using language models}.
\newblock In \emph{Proceedings of the 61st Annual Meeting of the Association
  for Computational Linguistics (Volume 1: Long Papers)}, pages 16477--16508,
  Toronto, Canada. Association for Computational Linguistics.

\bibitem[{Gao et~al.(2023{\natexlab{b}})Gao, Yen, Yu, and Chen}]{gao2023alce}
Tianyu Gao, Howard Yen, Jiatong Yu, and Danqi Chen. 2023{\natexlab{b}}.
\newblock \href {https://doi.org/10.18653/v1/2023.emnlp-main.398} {Enabling
  large language models to generate text with citations}.
\newblock In \emph{Proceedings of the 2023 Conference on Empirical Methods in
  Natural Language Processing}, pages 6465--6488, Singapore. Association for
  Computational Linguistics.

\bibitem[{Honovich et~al.(2022)Honovich, Aharoni, Herzig, Taitelbaum,
  Kukliansy, Cohen, Scialom, Szpektor, Hassidim, and Matias}]{honovich2022true}
Or~Honovich, Roee Aharoni, Jonathan Herzig, Hagai Taitelbaum, Doron Kukliansy,
  Vered Cohen, Thomas Scialom, Idan Szpektor, Avinatan Hassidim, and Yossi
  Matias. 2022.
\newblock \href {https://doi.org/10.18653/v1/2022.dialdoc-1.19} {{TRUE}:
  Re-evaluating factual consistency evaluation}.
\newblock In \emph{Proceedings of the Second DialDoc Workshop on
  Document-grounded Dialogue and Conversational Question Answering}, pages
  161--175, Dublin, Ireland. Association for Computational Linguistics.

\bibitem[{Hu et~al.(2022)Hu, Shen, Wallis, Allen{-}Zhu, Li, Wang, Wang, and
  Chen}]{hu2022lora}
Edward~J. Hu, Yelong Shen, Phillip Wallis, Zeyuan Allen{-}Zhu, Yuanzhi Li,
  Shean Wang, Lu~Wang, and Weizhu Chen. 2022.
\newblock \href {https://openreview.net/forum?id=nZeVKeeFYf9} {Lora: Low-rank
  adaptation of large language models}.
\newblock In \emph{The Tenth International Conference on Learning
  Representations, {ICLR} 2022, Virtual Event, April 25-29, 2022}.
  OpenReview.net.

\bibitem[{Huang et~al.(2024)Huang, Wu, Hu, and Wang}]{huang2024training}
Chengyu Huang, Zeqiu Wu, Yushi Hu, and Wenya Wang. 2024.
\newblock \href {https://arxiv.org/abs/2402.04315} {Training language models to
  generate text with citations via fine-grained rewards}.
\newblock \emph{ArXiv preprint}, abs/2402.04315.

\bibitem[{Huang et~al.(2021)Huang, Cao, Parulian, Ji, and
  Wang}]{huang-etal-2021-efficient}
Luyang Huang, Shuyang Cao, Nikolaus Parulian, Heng Ji, and Lu~Wang. 2021.
\newblock \href {https://doi.org/10.18653/v1/2021.naacl-main.112} {Efficient
  attentions for long document summarization}.
\newblock In \emph{Proceedings of the 2021 Conference of the North American
  Chapter of the Association for Computational Linguistics: Human Language
  Technologies}, pages 1419--1436, Online. Association for Computational
  Linguistics.

\bibitem[{Izacard et~al.(2021)Izacard, Caron, Hosseini, Riedel, Bojanowski,
  Joulin, and Grave}]{izacard2022unsupervised}
Gautier Izacard, Mathilde Caron, Lucas Hosseini, Sebastian Riedel, Piotr
  Bojanowski, Armand Joulin, and Edouard Grave. 2021.
\newblock \href {https://arxiv.org/abs/2112.09118} {Unsupervised dense
  information retrieval with contrastive learning}.
\newblock \emph{ArXiv preprint}, abs/2112.09118.

\bibitem[{Jiang et~al.(2023)Jiang, Sablayrolles, Mensch, Bamford, Chaplot,
  de~las Casas, Bressand, Lengyel, Lample, Saulnier, Lavaud, Lachaux, Stock,
  Scao, Lavril, Wang, Lacroix, and Sayed}]{jiang2023mistral}
Albert~Q. Jiang, Alexandre Sablayrolles, Arthur Mensch, Chris Bamford,
  Devendra~Singh Chaplot, Diego de~las Casas, Florian Bressand, Gianna Lengyel,
  Guillaume Lample, Lucile Saulnier, Lélio~Renard Lavaud, Marie-Anne Lachaux,
  Pierre Stock, Teven~Le Scao, Thibaut Lavril, Thomas Wang, Timothée Lacroix,
  and William~El Sayed. 2023.
\newblock \href {https://arxiv.org/abs/2310.06825} {Mistral 7b}.
\newblock \emph{ArXiv preprint}, abs/2310.06825.

\bibitem[{Kamoi et~al.(2023)Kamoi, Goyal, Diego~Rodriguez, and
  Durrett}]{kamoi-etal-2023-wice}
Ryo Kamoi, Tanya Goyal, Juan Diego~Rodriguez, and Greg Durrett. 2023.
\newblock \href {https://doi.org/10.18653/v1/2023.emnlp-main.470} {{W}i{CE}:
  Real-world entailment for claims in {W}ikipedia}.
\newblock In \emph{Proceedings of the 2023 Conference on Empirical Methods in
  Natural Language Processing}, pages 7561--7583, Singapore. Association for
  Computational Linguistics.

\bibitem[{Koreeda and Manning(2021)}]{koreeda-manning-2021-contractnli-dataset}
Yuta Koreeda and Christopher Manning. 2021.
\newblock \href {https://doi.org/10.18653/v1/2021.findings-emnlp.164}
  {{C}ontract{NLI}: A dataset for document-level natural language inference for
  contracts}.
\newblock In \emph{Findings of the Association for Computational Linguistics:
  EMNLP 2021}, pages 1907--1919, Punta Cana, Dominican Republic. Association
  for Computational Linguistics.

\bibitem[{Kuznetsov et~al.(2022)Kuznetsov, Buchmann, Eichler, and
  Gurevych}]{10.1162/coli_a_00455}
Ilia Kuznetsov, Jan Buchmann, Max Eichler, and Iryna Gurevych. 2022.
\newblock \href {https://doi.org/10.1162/coli_a_00455} {{Revise and Resubmit:
  An Intertextual Model of Text-based Collaboration in Peer Review}}.
\newblock \emph{Computational Linguistics}, 48(4):949--986.

\bibitem[{Kwan et~al.(2023)Kwan, Zeng, Wang, Sun, Li, Shang, Liu, and
  Wong}]{kwan2024m4le}
Wai-Chung Kwan, Xingshan Zeng, Yufei Wang, Yusen Sun, Liangyou Li, Lifeng
  Shang, Qun Liu, and Kam-Fai Wong. 2023.
\newblock \href {https://arxiv.org/abs/2310.19240} {M4le: A multi-ability
  multi-range multi-task multi-domain long-context evaluation benchmark for
  large language models}.
\newblock \emph{ArXiv preprint}, abs/2310.19240.

\bibitem[{Kwiatkowski et~al.(2019)Kwiatkowski, Palomaki, Redfield, Collins,
  Parikh, Alberti, Epstein, Polosukhin, Devlin, Lee, Toutanova, Jones, Kelcey,
  Chang, Dai, Uszkoreit, Le, and Petrov}]{kwiatkowski2019natural}
Tom Kwiatkowski, Jennimaria Palomaki, Olivia Redfield, Michael Collins, Ankur
  Parikh, Chris Alberti, Danielle Epstein, Illia Polosukhin, Jacob Devlin,
  Kenton Lee, Kristina Toutanova, Llion Jones, Matthew Kelcey, Ming-Wei Chang,
  Andrew~M. Dai, Jakob Uszkoreit, Quoc Le, and Slav Petrov. 2019.
\newblock \href {https://doi.org/10.1162/tacl_a_00276} {Natural questions: A
  benchmark for question answering research}.
\newblock \emph{Transactions of the Association for Computational Linguistics},
  7:452--466.

\bibitem[{Li et~al.(2023{\natexlab{a}})Li, Shao, Xie, Sheng, Zheng, Gonzalez,
  Stoica, and Zhang}]{longchat2023}
Dacheng Li, Rulin Shao, Anze Xie, Ying Sheng, Lianmin Zheng, Joseph~E.
  Gonzalez, Ion Stoica, and Hao Zhang. 2023{\natexlab{a}}.
\newblock \href {https://lmsys.org/blog/2023-06-29-longchat} {How long can
  open-source llms truly promise on context length?}
\newblock \emph{lmsys.org/blog}.

\bibitem[{Li et~al.(2023{\natexlab{b}})Li, Wang, Zheng, and
  Zhang}]{li2023loogle}
Jiaqi Li, Mengmeng Wang, Zilong Zheng, and Muhan Zhang. 2023{\natexlab{b}}.
\newblock \href {https://arxiv.org/abs/2311.04939} {Loogle: Can long-context
  language models understand long contexts?}
\newblock \emph{ArXiv preprint}, abs/2311.04939.

\bibitem[{Lin(2004)}]{lin-2004-rouge}
Chin-Yew Lin. 2004.
\newblock \href {https://aclanthology.org/W04-1013} {{ROUGE}: A package for
  automatic evaluation of summaries}.
\newblock In \emph{Text Summarization Branches Out}, pages 74--81, Barcelona,
  Spain. Association for Computational Linguistics.

\bibitem[{Lin et~al.(2023)Lin, Asai, Li, Oguz, Lin, Mehdad, tau Yih, and
  Chen}]{lin2023train}
Sheng-Chieh Lin, Akari Asai, Minghan Li, Barlas Oguz, Jimmy Lin, Yashar Mehdad,
  Wen tau Yih, and Xilun Chen. 2023.
\newblock \href {https://arxiv.org/abs/2302.07452} {How to train your dragon:
  Diverse augmentation towards generalizable dense retrieval}.
\newblock \emph{ArXiv preprint}, abs/2302.07452.

\bibitem[{Liu et~al.(2023)Liu, Zhang, and Liang}]{liu-etal-2023-evaluating}
Nelson Liu, Tianyi Zhang, and Percy Liang. 2023.
\newblock \href {https://doi.org/10.18653/v1/2023.findings-emnlp.467}
  {Evaluating verifiability in generative search engines}.
\newblock In \emph{Findings of the Association for Computational Linguistics:
  EMNLP 2023}, pages 7001--7025, Singapore. Association for Computational
  Linguistics.

\bibitem[{Liu et~al.(2024)Liu, Lin, Hewitt, Paranjape, Bevilacqua, Petroni, and
  Liang}]{liu2024lost}
Nelson~F. Liu, Kevin Lin, John Hewitt, Ashwin Paranjape, Michele Bevilacqua,
  Fabio Petroni, and Percy Liang. 2024.
\newblock \href {https://doi.org/10.1162/tacl_a_00638} {{Lost in the Middle:
  How Language Models Use Long Contexts}}.
\newblock \emph{Transactions of the Association for Computational Linguistics},
  12:157--173.

\bibitem[{Longpre et~al.(2023)Longpre, Hou, Vu, Webson, Chung, Tay, Zhou, Le,
  Zoph, Wei, and Roberts}]{pmlr-v202-longpre23a}
Shayne Longpre, Le~Hou, Tu~Vu, Albert Webson, Hyung~Won Chung, Yi~Tay, Denny
  Zhou, Quoc~V Le, Barret Zoph, Jason Wei, and Adam Roberts. 2023.
\newblock \href {https://proceedings.mlr.press/v202/longpre23a.html} {The flan
  collection: Designing data and methods for effective instruction tuning}.
\newblock In \emph{Proceedings of the 40th International Conference on Machine
  Learning}, volume 202 of \emph{Proceedings of Machine Learning Research},
  pages 22631--22648. PMLR.

\bibitem[{Loper and Bird(2002)}]{bird-2006-nltk}
Edward Loper and Steven Bird. 2002.
\newblock \href {https://doi.org/10.3115/1118108.1118117} {{NLTK}: The natural
  language toolkit}.
\newblock In \emph{Proceedings of the {ACL}-02 Workshop on Effective Tools and
  Methodologies for Teaching Natural Language Processing and Computational
  Linguistics}, pages 63--70, Philadelphia, Pennsylvania, USA. Association for
  Computational Linguistics.

\bibitem[{Loshchilov and Hutter(2019)}]{loshchilov2018decoupled}
Ilya Loshchilov and Frank Hutter. 2019.
\newblock \href {https://openreview.net/forum?id=Bkg6RiCqY7} {Decoupled weight
  decay regularization}.
\newblock In \emph{7th International Conference on Learning Representations,
  {ICLR} 2019, New Orleans, LA, USA, May 6-9, 2019}. OpenReview.net.

\bibitem[{Ma et~al.(2023)Ma, Zhang, Pradeep, and Lin}]{ma2023zeroshot}
Xueguang Ma, Xinyu Zhang, Ronak Pradeep, and Jimmy Lin. 2023.
\newblock \href {https://arxiv.org/abs/2305.02156} {Zero-shot listwise document
  reranking with a large language model}.
\newblock \emph{ArXiv preprint}, abs/2305.02156.

\bibitem[{Malaviya et~al.(2023)Malaviya, Lee, Chen, Sieber, Yatskar, and
  Roth}]{malaviya2024expertqa}
Chaitanya Malaviya, Subin Lee, Sihao Chen, Elizabeth Sieber, Mark Yatskar, and
  Dan Roth. 2023.
\newblock \href {https://arxiv.org/abs/2309.07852} {Expertqa: Expert-curated
  questions and attributed answers}.
\newblock \emph{ArXiv preprint}, abs/2309.07852.

\bibitem[{Muennighoff et~al.(2024)Muennighoff, Su, Wang, Yang, Wei, Yu, Singh,
  and Kiela}]{muennighoff2024generative}
Niklas Muennighoff, Hongjin Su, Liang Wang, Nan Yang, Furu Wei, Tao Yu,
  Amanpreet Singh, and Douwe Kiela. 2024.
\newblock \href {https://arxiv.org/abs/2402.09906} {Generative representational
  instruction tuning}.
\newblock \emph{ArXiv preprint}, abs/2402.09906.

\bibitem[{Ni et~al.(2022)Ni, Qu, Lu, Dai, Hernandez~Abrego, Ma, Zhao, Luan,
  Hall, Chang, and Yang}]{ni2021large}
Jianmo Ni, Chen Qu, Jing Lu, Zhuyun Dai, Gustavo Hernandez~Abrego, Ji~Ma,
  Vincent Zhao, Yi~Luan, Keith Hall, Ming-Wei Chang, and Yinfei Yang. 2022.
\newblock \href {https://aclanthology.org/2022.emnlp-main.669} {Large dual
  encoders are generalizable retrievers}.
\newblock In \emph{Proceedings of the 2022 Conference on Empirical Methods in
  Natural Language Processing}, pages 9844--9855, Abu Dhabi, United Arab
  Emirates. Association for Computational Linguistics.

\bibitem[{Rashkin et~al.(2023)Rashkin, Nikolaev, Lamm, Aroyo, Collins, Das,
  Petrov, Tomar, Turc, and Reitter}]{rashkin2023attribution}
Hannah Rashkin, Vitaly Nikolaev, Matthew Lamm, Lora Aroyo, Michael Collins,
  Dipanjan Das, Slav Petrov, Gaurav~Singh Tomar, Iulia Turc, and David Reitter.
  2023.
\newblock \href {https://doi.org/10.1162/coli_a_00486} {{Measuring Attribution
  in Natural Language Generation Models}}.
\newblock \emph{Computational Linguistics}, 49(4):777--840.

\bibitem[{Ravaut et~al.(2023)Ravaut, Sun, Chen, and Joty}]{ravaut2024context}
Mathieu Ravaut, Aixin Sun, Nancy~F. Chen, and Shafiq Joty. 2023.
\newblock \href {https://arxiv.org/abs/2310.10570} {On context utilization in
  summarization with large language models}.
\newblock \emph{ArXiv preprint}, abs/2310.10570.

\bibitem[{Reimers and Gurevych(2019)}]{reimers2019sentencebert}
Nils Reimers and Iryna Gurevych. 2019.
\newblock \href {https://doi.org/10.18653/v1/D19-1410} {Sentence-{BERT}:
  Sentence embeddings using {S}iamese {BERT}-networks}.
\newblock In \emph{Proceedings of the 2019 Conference on Empirical Methods in
  Natural Language Processing and the 9th International Joint Conference on
  Natural Language Processing (EMNLP-IJCNLP)}, pages 3982--3992, Hong Kong,
  China. Association for Computational Linguistics.

\bibitem[{Robertson and Zaragoza(2009)}]{robertson2009bm25}
Stephen Robertson and Hugo Zaragoza. 2009.
\newblock \href {https://doi.org/10.1561/1500000019} {The probabilistic
  relevance framework: Bm25 and beyond}.
\newblock \emph{Foundations and Trends® in Information Retrieval},
  3(4):333--389.

\bibitem[{Schimanski et~al.(2024)Schimanski, Ni, Kraus, Ash, and
  Leippold}]{schimanski2024faithful}
Tobias Schimanski, Jingwei Ni, Mathias Kraus, Elliott Ash, and Markus Leippold.
  2024.
\newblock \href {https://arxiv.org/abs/2402.08277} {Towards faithful and robust
  llm specialists for evidence-based question-answering}.
\newblock \emph{ArXiv preprint}, abs/2402.08277.

\bibitem[{Shaham et~al.(2023)Shaham, Ivgi, Efrat, Berant, and
  Levy}]{shaham-etal-2023-zeroscrolls}
Uri Shaham, Maor Ivgi, Avia Efrat, Jonathan Berant, and Omer Levy. 2023.
\newblock \href {https://doi.org/10.18653/v1/2023.findings-emnlp.536}
  {{Z}ero{SCROLLS}: A zero-shot benchmark for long text understanding}.
\newblock In \emph{Findings of the Association for Computational Linguistics:
  EMNLP 2023}, pages 7977--7989, Singapore. Association for Computational
  Linguistics.

\bibitem[{Slobodkin et~al.(2023)Slobodkin, Goldman, Caciularu, Dagan, and
  Ravfogel}]{slobodkin2023curious}
Aviv Slobodkin, Omer Goldman, Avi Caciularu, Ido Dagan, and Shauli Ravfogel.
  2023.
\newblock \href {https://doi.org/10.18653/v1/2023.emnlp-main.220} {The curious
  case of hallucinatory (un)answerability: Finding truths in the hidden states
  of over-confident large language models}.
\newblock In \emph{Proceedings of the 2023 Conference on Empirical Methods in
  Natural Language Processing}, pages 3607--3625, Singapore. Association for
  Computational Linguistics.

\bibitem[{Staniszewski et~al.(2023)Staniszewski, Tworkowski, Zhao, Jaszczur,
  Michalewski, Łukasz Kuciński, and Miłoś}]{staniszewski2024structured}
Konrad Staniszewski, Szymon Tworkowski, Yu~Zhao, Sebastian Jaszczur, Henryk
  Michalewski, Łukasz Kuciński, and Piotr Miłoś. 2023.
\newblock \href {https://arxiv.org/abs/2312.17296} {Structured packing in llm
  training improves long context utilization}.
\newblock \emph{ArXiv preprint}, abs/2312.17296.

\bibitem[{Sun et~al.(2023)Sun, Liu, Wang, Zhu, and Iyyer}]{sun2023pearl}
Simeng Sun, Yang Liu, Shuohang Wang, Chenguang Zhu, and Mohit Iyyer. 2023.
\newblock \href {https://arxiv.org/abs/2305.14564} {Pearl: Prompting large
  language models to plan and execute actions over long documents}.
\newblock \emph{ArXiv preprint}, abs/2305.14564.

\bibitem[{Tang et~al.(2024)Tang, Laban, and Durrett}]{tang2024minicheck}
Liyan Tang, Philippe Laban, and Greg Durrett. 2024.
\newblock \href {https://arxiv.org/abs/2404.10774} {Minicheck: Efficient
  fact-checking of llms on grounding documents}.
\newblock \emph{ArXiv preprint}, abs/2404.10774.

\bibitem[{Team et~al.(2024)Team, Mesnard, Hardin, Dadashi, Bhupatiraju, Pathak,
  Sifre, Rivière, Kale, Love, Tafti, Hussenot, Sessa, Chowdhery, Roberts,
  Barua, Botev, Castro-Ros, Slone, Héliou, Tacchetti, Bulanova, Paterson,
  Tsai, Shahriari, Lan, Choquette-Choo, Crepy, Cer, Ippolito, Reid,
  Buchatskaya, Ni, Noland, Yan, Tucker, Muraru, Rozhdestvenskiy, Michalewski,
  Tenney, Grishchenko, Austin, Keeling, Labanowski, Lespiau, Stanway, Brennan,
  Chen, Ferret, Chiu, Mao-Jones, Lee, Yu, Millican, Sjoesund, Lee, Dixon, Reid,
  Mikuła, Wirth, Sharman, Chinaev, Thain, Bachem, Chang, Wahltinez, Bailey,
  Michel, Yotov, Chaabouni, Comanescu, Jana, Anil, McIlroy, Liu, Mullins,
  Smith, Borgeaud, Girgin, Douglas, Pandya, Shakeri, De, Klimenko, Hennigan,
  Feinberg, Stokowiec, hui Chen, Ahmed, Gong, Warkentin, Peran, Giang, Farabet,
  Vinyals, Dean, Kavukcuoglu, Hassabis, Ghahramani, Eck, Barral, Pereira,
  Collins, Joulin, Fiedel, Senter, Andreev, and Kenealy}]{gemmateam2024gemma}
Gemma Team, Thomas Mesnard, Cassidy Hardin, Robert Dadashi, Surya Bhupatiraju,
  Shreya Pathak, Laurent Sifre, Morgane Rivière, Mihir~Sanjay Kale, Juliette
  Love, Pouya Tafti, Léonard Hussenot, Pier~Giuseppe Sessa, Aakanksha
  Chowdhery, Adam Roberts, Aditya Barua, Alex Botev, Alex Castro-Ros, Ambrose
  Slone, Amélie Héliou, Andrea Tacchetti, Anna Bulanova, Antonia Paterson,
  Beth Tsai, Bobak Shahriari, Charline~Le Lan, Christopher~A. Choquette-Choo,
  Clément Crepy, Daniel Cer, Daphne Ippolito, David Reid, Elena Buchatskaya,
  Eric Ni, Eric Noland, Geng Yan, George Tucker, George-Christian Muraru,
  Grigory Rozhdestvenskiy, Henryk Michalewski, Ian Tenney, Ivan Grishchenko,
  Jacob Austin, James Keeling, Jane Labanowski, Jean-Baptiste Lespiau, Jeff
  Stanway, Jenny Brennan, Jeremy Chen, Johan Ferret, Justin Chiu, Justin
  Mao-Jones, Katherine Lee, Kathy Yu, Katie Millican, Lars~Lowe Sjoesund, Lisa
  Lee, Lucas Dixon, Machel Reid, Maciej Mikuła, Mateo Wirth, Michael Sharman,
  Nikolai Chinaev, Nithum Thain, Olivier Bachem, Oscar Chang, Oscar Wahltinez,
  Paige Bailey, Paul Michel, Petko Yotov, Rahma Chaabouni, Ramona Comanescu,
  Reena Jana, Rohan Anil, Ross McIlroy, Ruibo Liu, Ryan Mullins, Samuel~L
  Smith, Sebastian Borgeaud, Sertan Girgin, Sholto Douglas, Shree Pandya,
  Siamak Shakeri, Soham De, Ted Klimenko, Tom Hennigan, Vlad Feinberg, Wojciech
  Stokowiec, Yu~hui Chen, Zafarali Ahmed, Zhitao Gong, Tris Warkentin, Ludovic
  Peran, Minh Giang, Clément Farabet, Oriol Vinyals, Jeff Dean, Koray
  Kavukcuoglu, Demis Hassabis, Zoubin Ghahramani, Douglas Eck, Joelle Barral,
  Fernando Pereira, Eli Collins, Armand Joulin, Noah Fiedel, Evan Senter, Alek
  Andreev, and Kathleen Kenealy. 2024.
\newblock \href {https://arxiv.org/abs/2403.08295} {Gemma: Open models based on
  gemini research and technology}.
\newblock \emph{ArXiv preprint}, abs/2403.08295.

\bibitem[{Thakur et~al.(2021)Thakur, Reimers, R{\"u}ckl{\'e}, Srivastava, and
  Gurevych}]{thakur2021beir}
Nandan Thakur, Nils Reimers, Andreas R{\"u}ckl{\'e}, Abhishek Srivastava, and
  Iryna Gurevych. 2021.
\newblock \href {https://openreview.net/forum?id=wCu6T5xFjeJ} {{BEIR}: A
  heterogeneous benchmark for zero-shot evaluation of information retrieval
  models}.
\newblock In \emph{Thirty-fifth Conference on Neural Information Processing
  Systems Datasets and Benchmarks Track (Round 2)}.

\bibitem[{Touvron et~al.(2023)Touvron, Lavril, Izacard, Martinet, Lachaux,
  Lacroix, Rozière, Goyal, Hambro, Azhar, Rodriguez, Joulin, Grave, and
  Lample}]{touvron2023llama}
Hugo Touvron, Thibaut Lavril, Gautier Izacard, Xavier Martinet, Marie-Anne
  Lachaux, Timothée Lacroix, Baptiste Rozière, Naman Goyal, Eric Hambro,
  Faisal Azhar, Aurelien Rodriguez, Armand Joulin, Edouard Grave, and Guillaume
  Lample. 2023.
\newblock \href {https://arxiv.org/abs/2302.13971} {Llama: Open and efficient
  foundation language models}.

\bibitem[{Wolf et~al.(2020)Wolf, Debut, Sanh, Chaumond, Delangue, Moi, Cistac,
  Rault, Louf, Funtowicz, Davison, Shleifer, von Platen, Ma, Jernite, Plu, Xu,
  Le~Scao, Gugger, Drame, Lhoest, and Rush}]{wolf2020transformers}
Thomas Wolf, Lysandre Debut, Victor Sanh, Julien Chaumond, Clement Delangue,
  Anthony Moi, Pierric Cistac, Tim Rault, Remi Louf, Morgan Funtowicz, Joe
  Davison, Sam Shleifer, Patrick von Platen, Clara Ma, Yacine Jernite, Julien
  Plu, Canwen Xu, Teven Le~Scao, Sylvain Gugger, Mariama Drame, Quentin Lhoest,
  and Alexander Rush. 2020.
\newblock \href {https://doi.org/10.18653/v1/2020.emnlp-demos.6} {Transformers:
  State-of-the-art natural language processing}.
\newblock In \emph{Proceedings of the 2020 Conference on Empirical Methods in
  Natural Language Processing: System Demonstrations}, pages 38--45, Online.
  Association for Computational Linguistics.

\bibitem[{Wu et~al.(2024)Wu, Iso, Pezeshkpour, Bhutani, and
  Hruschka}]{wu-etal-2024-less}
Yunshu Wu, Hayate Iso, Pouya Pezeshkpour, Nikita Bhutani, and Estevam Hruschka.
  2024.
\newblock \href {https://aclanthology.org/2024.eacl-short.29} {Less is more for
  long document summary evaluation by {LLM}s}.
\newblock In \emph{Proceedings of the 18th Conference of the European Chapter
  of the Association for Computational Linguistics (Volume 2: Short Papers)},
  pages 330--343, St. Julian{'}s, Malta. Association for Computational
  Linguistics.

\bibitem[{Xu et~al.(2024)Xu, Ping, Wu, McAfee, Zhu, Liu, Subramanian,
  Bakhturina, Shoeybi, and Catanzaro}]{xu2024retrieval}
Peng Xu, Wei Ping, Xianchao Wu, Lawrence McAfee, Chen Zhu, Zihan Liu, Sandeep
  Subramanian, Evelina Bakhturina, Mohammad Shoeybi, and Bryan Catanzaro. 2024.
\newblock \href {https://openreview.net/forum?id=xw5nxFWMlo} {Retrieval meets
  long context large language models}.
\newblock In \emph{The Twelfth International Conference on Learning
  Representations}.

\bibitem[{Yue et~al.(2023)Yue, Wang, Chen, Zhang, Su, and
  Sun}]{yue-etal-2023-automatic}
Xiang Yue, Boshi Wang, Ziru Chen, Kai Zhang, Yu~Su, and Huan Sun. 2023.
\newblock \href {https://doi.org/10.18653/v1/2023.findings-emnlp.307}
  {Automatic evaluation of attribution by large language models}.
\newblock In \emph{Findings of the Association for Computational Linguistics:
  EMNLP 2023}, pages 4615--4635, Singapore. Association for Computational
  Linguistics.

\bibitem[{Zhang et~al.(2020{\natexlab{a}})Zhang, Zhao, Saleh, and
  Liu}]{zhang2020pegasus}
Jingqing Zhang, Yao Zhao, Mohammad Saleh, and Peter~J. Liu. 2020{\natexlab{a}}.
\newblock \href {http://proceedings.mlr.press/v119/zhang20ae.html} {{PEGASUS:}
  pre-training with extracted gap-sentences for abstractive summarization}.
\newblock In \emph{Proceedings of the 37th International Conference on Machine
  Learning, {ICML} 2020, 13-18 July 2020, Virtual Event}, volume 119 of
  \emph{Proceedings of Machine Learning Research}, pages 11328--11339. {PMLR}.

\bibitem[{Zhang et~al.(2020{\natexlab{b}})Zhang, Kishore, Wu, Weinberger, and
  Artzi}]{Zhang*2020BERTScore:}
Tianyi Zhang, Varsha Kishore, Felix Wu, Kilian~Q. Weinberger, and Yoav Artzi.
  2020{\natexlab{b}}.
\newblock \href {https://openreview.net/forum?id=SkeHuCVFDr} {Bertscore:
  Evaluating text generation with {BERT}}.
\newblock In \emph{8th International Conference on Learning Representations,
  {ICLR} 2020, Addis Ababa, Ethiopia, April 26-30, 2020}. OpenReview.net.

\end{thebibliography}

\appendix

\section{Additional Results}
\label{sec:appendix/additional_results}
\subsection{Correlation between Evidence Position and Response Quality}

See Table~\ref{tab:evidence_position_performance}.

\begin{table*}[]
    \small
    \centering
    \begin{tabular}{ll|cccccc}
\toprule
 & & QSP & NQ & EI & WIC & CNLI & GR \\
\midrule
\multirow[t]{2}{*}{GPT-3.5} & post-hoc & -0.76* & -0.03 & -0.77* & -0.49 & 0.42 & -0.45 \\
 & citation & -0.08 & -0.44 & -0.62 & -0.41 & -0.02 & -0.59 \\
\cline{1-8}
\multirow[t]{2}{*}{GPT-4} & post-hoc & 0.12 & 0.02 & 0.03 & 0.35 & 0.04 & 0.12 \\
 & citation & 0.12 & 0.04 & 0.03 & 0.09 & -0.19 & 0.03 \\
\cline{1-8}
\multirow[t]{2}{*}{Flan-T5} & post-hoc & -0.32 & 0.05 & -0.75* & -0.37 & -0.07 & -0.92* \\
 & citation & -0.19 & -0.13 & -0.81* & -0.76* & -0.26 & -0.95* \\
\cline{1-8}
\multirow[t]{2}{*}{Longchat} & post-hoc & -0.43 & 0.44 & -0.33 & -0.32 & -0.58 & -0.64* \\
 & citation & -0.49 & -0.28 & -0.42 & -0.30 & -0.54 & -0.61 \\
\cline{1-8}
\multirow[t]{2}{*}{Mistral} & post-hoc & -0.53 & 0.39 & -0.59 & 0.04 & 0.09 & -0.92* \\
 & citation & -0.35 & 0.31 & -0.37 & -0.58 & 0.47 & -0.82* \\
\cline{1-8}
\bottomrule
\end{tabular}
    \caption{Pearson correlation between response quality and position of annotated evidence. * indicates significance ($p<0.05$). See §\ref{sec:experiments/positional_biases} for details.}
    \label{tab:evidence_position_performance}
\end{table*}

\subsection{System-level Correlation of Attributability and Response Quality}

\begin{table}[]
    \centering
    \small
\begin{tabular}{l|c}
\toprule
 & Pearson $r$ ($p$) \\
\midrule
QSP & $0.93$ ($7.7 \times 10^{-5}$) \\
NQ & $0.99$ ($3.6 \times 10^{-8}$) \\
EI & $0.93$ ($7.9 \times 10^{-5}$) \\
WIC & $0.96$ ($1.2 \times 10^{-5}$) \\
CNLI & $0.93$ ($1.2 \times 10^{-4}$) \\
GR & $0.36$ ($3.0 \times 10^{-1}$) \\
mean & $0.98$ ($1.3 \times 10^{-6}$) \\
\bottomrule
\end{tabular}
    \caption{System-level Pearson correlation between response quality and evidence quality for \texttt{citation} and \texttt{reduced-citation} (8 score pairs per dataset). For Avg, correlation was computed over the average scores (rightmost columns in Table~\ref{tab:main_table}).}
    \label{tab:correlation}
\end{table}

See Table~\ref{tab:correlation}

\subsection{Overlap between \texttt{citation} and \texttt{post-hoc} evidence}

\subsection{Relation between Performance and Input Document Length}

See Tables \ref{tab:response_quality_vs_doc_length_correlation} and \ref{tab:evidence_quality_vs_doc_length_correlation}. See §\ref{sec:experiments/optimal_methods_for_attribution} for analysis and discussion.

\begin{table*}[]
    \small
    \centering
    \begin{tabular}{ll|cccccc}
\toprule
 & & QSP & NQ & EI & WIC & CNLI & GR \\
\midrule
\multirow[t]{2}{*}{GPT-3.5} & citation & 0.04 & -0.01 & -0.08* & -0.04 & 0.05* & -0.08* \\
 & post-hoc & 0.08* & -0.00 & -0.05 & -0.05 & 0.05* & -0.04 \\
\cline{1-8}
\multirow[t]{2}{*}{GPT-4} & citation & 0.29* & 0.11 & 0.13 & 0.08 & 0.34* & -0.34* \\
 & post-hoc & 0.28* & 0.17 & 0.13 & 0.11 & 0.39* & -0.26* \\
\cline{1-8}
\multirow[t]{2}{*}{Flan-T5} & citation & -0.02 & -0.10* & -0.07* & 0.05 & 0.00 & -0.07* \\
 & post-hoc & -0.03 & -0.10* & -0.05 & -0.08 & -0.02 & -0.11* \\
\cline{1-8}
\multirow[t]{2}{*}{Longchat} & citation & -0.05 & -0.05* & -0.04 & 0.01 & 0.04 & -0.15* \\
 & post-hoc & -0.02 & -0.13* & -0.04 & 0.03 & -0.06* & -0.13* \\
\cline{1-8}
\multirow[t]{2}{*}{Mistral} & citation & -0.03 & -0.07* & -0.07* & -0.07 & 0.00 & -0.12* \\
 & post-hoc & -0.01 & -0.06* & -0.07* & -0.04 & -0.02 & -0.06 \\
\bottomrule
\end{tabular}
    \caption{Pearson correlation between response quality and document length. * indicates significance ($p<0.05$). See §\ref{sec:experiments/optimal_methods_for_attribution} for analysis and discussion.}
    \label{tab:response_quality_vs_doc_length_correlation}
\end{table*}

\begin{table*}[]
    \small
    \centering
    \begin{tabular}{ll|cccccc}
\toprule
 & & QSP & NQ & EI & WIC & CNLI & GR \\
\midrule
\multirow[t]{2}{*}{GPT-3.5} & citation & -0.08* & -0.13* & -0.10* & -0.20* & -0.04 & -0.24* \\
 & post-hoc & -0.03 & -0.10* & -0.01 & -0.11* & -0.07* & -0.23* \\
\cline{1-8}
\multirow[t]{2}{*}{GPT-4} & citation & 0.04 & -0.17 & -0.09 & 0.09 & 0.15 & -0.10 \\
 & post-hoc & -0.13 & -0.05 & -0.07 & 0.10 & 0.01 & -0.12 \\
\cline{1-8}
\multirow[t]{2}{*}{Flan-T5} & citation & -0.04 & -0.08 & -0.10* & -0.09 & -0.07* & -0.11* \\
 & post-hoc & -0.06* & -0.20* & -0.00 & -0.15* & -0.13* & -0.10* \\
\cline{1-8}
\multirow[t]{2}{*}{Longchat} & citation & -0.13* & -0.09* & -0.12* & -0.05 & 0.08* & -0.03 \\
 & post-hoc & -0.01 & -0.13* & 0.00 & 0.03 & 0.04* & -0.18* \\
\cline{1-8}
\multirow[t]{2}{*}{Mistral} & citation & -0.18* & -0.17* & -0.13* & -0.19* & -0.09* & -0.05 \\
 & post-hoc & -0.08* & -0.10* & -0.04 & -0.16* & -0.06* & -0.10* \\
\bottomrule
\end{tabular}
    \caption{Pearson correlation between evidence quality and document length. * indicates significance ($p<0.05$). See §\ref{sec:experiments/optimal_methods_for_attribution} for analysis and discussion.}
    \label{tab:evidence_quality_vs_doc_length_correlation}
\end{table*}

\subsection{Effect of Evidence Position on Evidence Quality}
\label{appendix:evidence_position_evidence_quality}

\begin{figure*}
    \centering
    \includegraphics{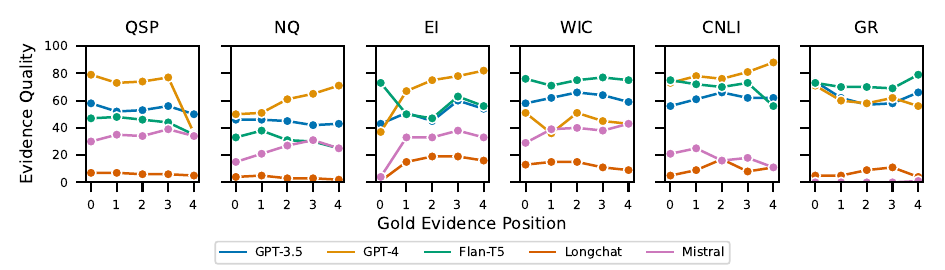}
    \caption{Evidence quality by position of gold evidence in the document. ``Lost in the Middle'' effect is not visible. For more information, see §\ref{appendix:evidence_position_evidence_quality}.}
    \label{fig:gold_evidence_position_evidence_quality}
\end{figure*}

\begin{figure*}
    \centering
    \includegraphics{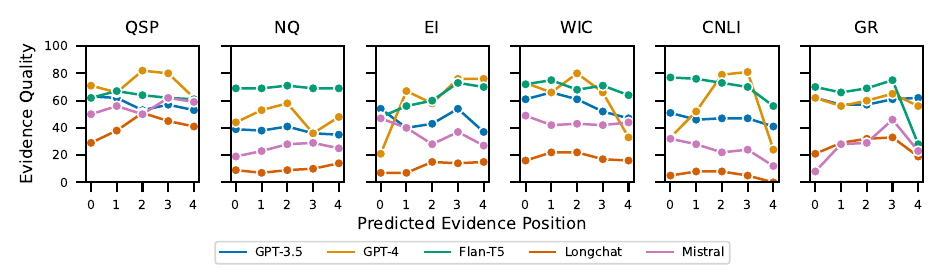}
    \caption{Evidence quality by position of predicted evidence in the document. A ``Lost in the Middle'' effect is not visible. For more information, see §\ref{appendix:evidence_position_evidence_quality}.}
    \label{fig:predicted_evidence_position_evidence_quality}
\end{figure*}

To analyze the effect of the position of gold or predicted evidence on evidence quality, we grouped predictions from models under the \texttt{citation} approach into 5 bins based on the relative position of gold or predicted evidence in the document. We then computed mean evidence quality separately for each bin. Figures \ref{fig:gold_evidence_position_evidence_quality} and \ref{fig:predicted_evidence_position_evidence_quality} show the results. A ``Lost in the Middle'' effect is not visible.

\section{Datasets}
\label{sec:appendix/datasets}

\begin{table}[]
\small
\centering
\begin{tabularx}{7cm}{l|X}
\toprule
QASPER             & CC-BY 4.0                                 \\
Natural Questions  & CC-BY-SA 3.0                              \\
Evidence Inference & MIT                                       \\
Wice               & CC-BY-SA 4.0 (text), ODC-BY (annotations) \\
ContractNLI        & CC-BY 4.0                                 \\
GovReport          & No copyright    \\
\bottomrule
\end{tabularx}
\caption{Licenses of the datasets in \benchmarkname{}.}
\label{tab:dataset_licenses}
\end{table}

See Table~\ref{tab:dataset_examples} for examples from the datasets in \benchmarkname{}

\paragraph{QASPER} \cite{dasigi-etal-2021-dataset} is a dataset of NLP papers and questions about them. Answers can be extractive, abstractive, ``Yes'', ``No'' or ``unanswerable''. Evidence is annotated on paragraph level. We remove instances with evidence in tables or figures.

\paragraph{Natural Questions} \cite{kwiatkowski2019natural} is a dataset of genuine questions from Google search logs and Wikipedia pages that may or may not contain the answers. We removed all annotations with answers in tables and those that only have a \textit{long} answer, keeping only the annotations that have short answers (i.e. extractive spans), ``Yes''/``No'' answers or ``unanswerable''. All non-unanswerable annotations have a single evidence paragraph. As the official test set is hidden, we used the dev set for testing and a part of the train set for development. 

\paragraph{Evidence Inference} \cite{deyoung-etal-2020-evidence} consists of reports from clinical studies, "prompts" in the form of \textit{intervention}, \textit{comparator}, and \textit{outcome}, one or multiple labels for the prompt ("significantly increased", "significantly decreased", or "no significant difference") and corresponding evidence spans. We map the annotated evidence spans to paragraphs. 

\paragraph{Wice} \cite{kamoi-etal-2023-wice} is a dataset of claims from Wikipedia and referenced webpages. Claims are annotated as ``supported'', ``partially supported'' or ``not supported''. The referenced webpages are annotated with evidence on sentence level. We use the full-claim subset. 

\paragraph{ContractNLI} \cite{koreeda-manning-2021-contractnli-dataset} is a dataset of non-disclosure agreement contracts and claims about these agreements. The relation between contract and claim is annotated as ``entailment'', ``contradiction'' or ``not mentioned''. We split the contract documents into paragraphs at newline symbols to obtain paragraphs, and map the sentence-level annotated evidence to these paragraphs. 

\paragraph{GovReport} \cite{huang-etal-2021-efficient} is a dataset of reports from US-American governmental institutions and their executive summaries. 

\paragraph{Dataset format} All datasets were converted to the Intertext Graph format \cite{10.1162/coli_a_00455} to enable shared processing and the use of document structure (where available). 

\section{Prompts}
\label{sec:appendix/prompts}
The prompts used in our experiments can be divided into two building blocks: (1) Instructioncand (2) instance specific input. The instruction further consists of (a) task explanation and (b) format explanation. We explain the building blocks in the following. For a complete prompt example, see Table~\ref{tab:full_prompt_example}.

\subsection{Instruction}

\subsubsection{Task explanation}

Task explanations give information on the type of document used as input, the type of task to be solved, and possible labels. Table~\ref{tab:task_explanations} shows the task explanations used.

\subsubsection{Format explanation}

Format explanations were only used for \texttt{citation} approaches, explaining the expected format of pointers to evidence segments. Table~\ref{tab:format_explanations} shows the format explanations used for single-statement responses and multi-statement responses. The single statement explanation was used for all datasets except GovReport, where the multi statement explanation was used. 

\subsection{Instance Specific Input}

Instance specific input consisted of an input document and task-dependent additional information, such as a question or a claim. Table~\ref{tab:instance_specific_input} shows the formatting of instance specific input. 

\subsection{Example document formatting} We shortened the documents in examples to document title, section headings (where available) and annotated evidence segments. If there were no annotated evidence segments (e.g. because an example instance is unanswerable) we selected 2 random segments from the document (5 for Wice). 

\subsection{Prompt Selection} 

We optimized two prompt properties: The position of the instruction (i.e. task explanation and format explanation) and the number of few-shot examples. We ran experiments employing GPT-3.5 on QASPER and GovReport under the \texttt{citation} approach, the results are shown in Table \ref{tab:prompt_optimization}. We first varied the position of the instruction, finding that the instruction before the instance specific input resulted in best performance. Next, we experimented with using 1, 2 or 3 few shot examples, finding that 3 examples resulted in best performance. We limited the number of few-shot examples to leave enough space for the input document.

\subsection{Complete prompt example}

See Table~\ref{tab:full_prompt_example}.

\section{Attributability Evaluation}
\label{sec:appendix/attributability_evaluation}
To evaluate attributability, we experimented with TRUE \cite{honovich2022true}, Attrscore (Flan-T5-XXL version) \cite{yue-etal-2023-automatic} and Minicheck (Flan-T5-Large version) \cite{tang2024minicheck}. These models expect a \textit{claim} and \textit{evidence} as input. In the following we explain the construction of claims, evidence formatting and model-specific prompts. 

\subsection{Claim Construction} 

Claims were constructed based on the task specific inputs and outputs. Table \ref{tab:claim_examples} shows examples.

\paragraph{QA datasets} For QASPER and Natural Questions, question and answer were concatenated to get the claim.

\paragraph{Evidence Inference} If the response was ``no significant difference'', the claim was formulated as ``There was no significant difference between the effect of \{intervention\} and the effect of \{comparator\} on \{outcome\}.''. If the response was ``significantly increased'' or ``significantly decreased'' were predicted, the claim was formulated as ``The \{intervention\} \{response\} \{outcome\} in comparison to \{comparator\}''.

\paragraph{Wice} If the response was ``supported'', the input claim was used as the claim for attributability evaluation. If the response was ``partially supported'', the claim for attributability evaluation was formulated as ``The claim \{claim\} is partially supported.'' ``Not supported'' responses did not require attributability evaluation.

\paragraph{ContractNLI} If the response was ``entailment'', the input claim was used as the claim for attributability evaluation. As there are only 20 claims in the complete dataset, we formulated an \textit{inverse} version of each claim. This was used as the claim when the response was ``contradiction''. See Table~\ref{tab:claim_examples} for an example.

\paragraph{GovReport} The generated summary sentences were used as claims. 

\subsection{Input Formatting}

\paragraph{Evidence formatting} Predicted evidence segments were ordered by occurence in the document, joined by newline symbols, and prepended with the document title.

\paragraph{Model-specific formatting}

The prompt templates for attributability evaluation can be seen in Table~\ref{tab:attributability_evaluation_prompts}. They were taken from the respective original publications. 

\subsection{Annotation Instructions}

\begin{figure*}
    \centering
    \includegraphics[page=2, width=\textwidth]{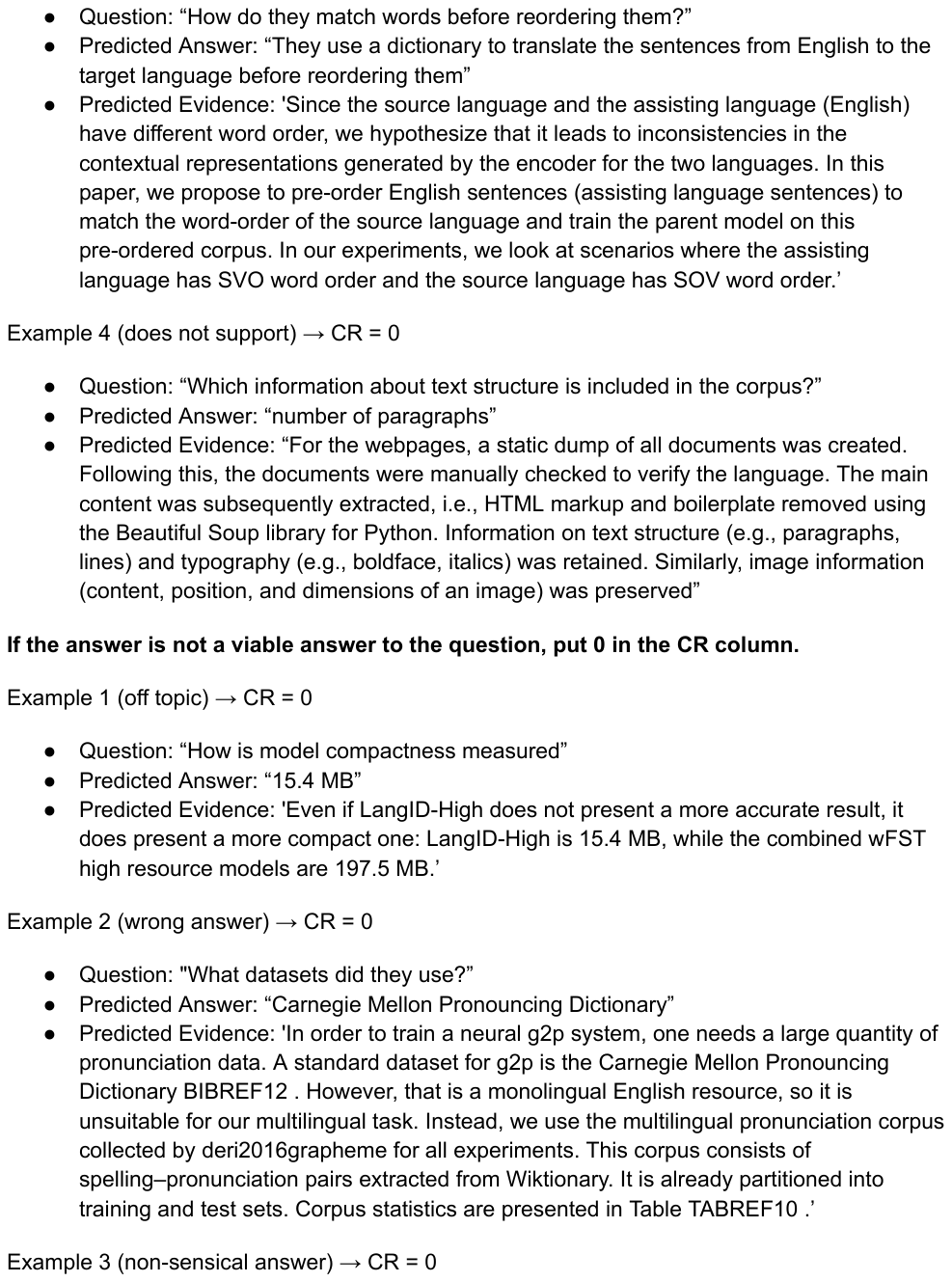}
    \caption{Annotation instructions for attributability model evaluation.}
    \label{fig:instructions}
\end{figure*}

As mentioned in \ref{sec:methods/evaluation/attributability_eval}, we manually annotated predictions on QASPER, Natural Questions and GovReport. The annotation instructions are shown in Fig.~\ref{fig:instructions}.

\section{Generation}
\label{sec:appendix/generation_model}
\subsection{Model Selection}

To select open source models for prompting and fine-tuning, we compared their performance in preliminary experiments. Table~\ref{tab:candidate_models} shows all open source models considered. 

\paragraph{Fine-tuning} We fine-tuned all candidate models for 1000 steps on QASPER and evaluated them on the dev set. As the results in Table~\ref{tab:finetune_model_selection} show that Flan-T5-XL has a clear advantage over the other models, we used it in all further fine-tuning experiments.

\paragraph{Prompting} We evaluated all candidate models on 100 instances from the QASPER and GovReport dev sets, using the \texttt{citation} approach with the prompts described in §\ref{sec:appendix/prompts}. Table~\ref{tab:prompt_model_selection} shows the results. As the Longchat model obtained the highest average score, we used it in all further experiments.

\subsection{Hyperparameters}

\paragraph{Generation} We set the maximum input length to 16K, truncating the input document if needed. We performed greedy decoding and temperature 0 for best reproducibility (§\ref{sec:appendix/technical_details}).

\paragraph{Fine-tuning} We employed LoRA fine-tuning \cite{hu2022lora} in a \texttt{citation} setting and a non-\texttt{citation} setting for 1000 steps. We set $r=64$, $\alpha=16$, a dropout rate of $0.1$, a learning rate of $10^-4$, effective batch size of $8$, and employed an AdamW optimizer \cite{loshchilov2018decoupled}

\section{Retrieval}
\label{sec:appendix/retrieval}
\subsection{Retriever Selection}

\paragraph{Candidates} We experimented with BM25 \cite{robertson2009bm25}, GTR \cite{ni2021large}, Contriever \cite{izacard2022unsupervised}, Dragon \cite{lin2023train} and the best-performing Sentence Transformer ``all-mpnet-base-v2''\footnote{\url{https://sbert.net/docs/sentence_transformer/pretrained_models.html\#semantic-search-models}} \cite{reimers2019sentencebert}.

\paragraph{Results} We tested all combinations of \texttt{post-hoc} (Table~\ref{tab:post_hoc_retriever_selection}), \texttt{retrieve-then-read} (Table~\ref{tab:rtr_retriever_selection}) and \texttt{reduced-citation} (Table~\ref{tab:reduce_retriever}) approaches, tasks and retrievers, using GPT-3.5 to generate responses. We selected the best-performing retriever for each combination

\subsection{Query Construction}

\paragraph{{\texttt{Post-hoc} query construction}} \texttt{Post-hoc} queries were constructed by combining instance specific inputs and outputs. The exact query construction depended on the task. For QASPER and Natural Questions, question and generated response were concatenated. For Evidence Inference, ContractNLI and GovReport, \texttt{post-hoc} queries were constructed in the same manner as claims for attributability evaluation (see §\ref{sec:appendix/attributability_evaluation}). For Wice, the input claim was used as the query. See Table~\ref{tab:post_hoc_query_examples} for examples.

\paragraph{\texttt{Reduce} and \texttt{retrieve-then-read} query construction} Queries for \texttt{Reduce} and \texttt{retrieve-then-read} were constructed based on instance specific input. For QASPER and Natural Questions, this was the question. For Evidence Inference, this was the question formed out of intervention, comparator and outcome. For Wice and ContractNLI, this was the claim. For GovReport, these were the document paragraphs. See Table~\ref{tab:retrieve-then-read_query_examples} for examples.

\paragraph{\texttt{Reduce} and \texttt{retrieve-then-read} for GovReport} To find the most relevant paragraphs from documents in the GovReport dataset, we used each paragraph as a query and computed the retrieval score for all paragraphs (including the paragraph itself), resulting in $n^2$ scores $s_{i,j}$ for a document with $n$ paragraphs. Each $s_{i,j}$ is the score for retrieving $p_j$ with query $p_i$. We compute a single score for each paragraph as $ s_j^* = \sum_{i=0}^{n} s_{i,j}$, i.e. the sum of scores to retrieve $p_j$. The paragraphs with the highest $s^*$ are then selected.

\section{Technical Details}
\label{sec:appendix/technical_details}
\paragraph{Technical setup} GPT-35 and GPT-4 were accessed via the Azure OpenAI API \footnote{\url{https://azure.microsoft.com/en-us/products/ai-services/openai-service}}, all other models were downloaded and run locally via Huggingface Transformers \cite{wolf2020transformers} on NVIDIA A100 and H100 GPUs.

\paragraph{Rouge Scoring} We used the \texttt{rouge-score} package\footnote{\url{https://pypi.org/project/rouge-score/}} to evaluate ROUGE-L

\paragraph{Compute Budget} We spent around \$400 to access OpenAI models, including preliminary experiments. We estimate to have spent 300 GPU hours on all experiments, including fine tuning, inference and attributability evaluation. 

\paragraph{Use of AI assistants} We used Github Copilot\footnote{\url{https://github.com/features/copilot}} for coding assistance. 

\paragraph{Detecting abstaining} 

Based on \cite{slobodkin2023curious} and our own inspection of model responses, we compiled two sets of keyphrases to detect abstaining. If any of these key was found in a response, it was set to ``unanswerable''. The first set of keyphrases is shown in Table~\ref{tab:unanswerable_keywords}, and was used as is. The second set of keyphrases was constructed by combining all verbs and adverbs in Table~\ref{tab:unanswerable_phrases} as follows: 
\begin{enumerate}
    \item ``not <verb>''
    \item ``not <adverb> <verb>''
\end{enumerate}
Verbs were used in their base form and as participles (e.g. ``mention'' $\rightarrow$ ``mentioned''). This means that each combination of verb and adverb results in four keyphrases. For example, the verb ``provide'' and the adverb ``explicitly'' resulted in ``not provide'', ``not explicitly provide'', ``not provided'', ``not explicitly provided''.

\begin{table*}[]
\small
\small
\begin{tabularx}{\textwidth}{l|X|X}
\toprule
                   & Input                                                                                                                                                                         & Output                                                              \\
\midrule
QASPER             & Question: Which domains do they explore?                                                                                                                                                & news articles related to Islam and articles discussing Islam basics \\ \hline
Natural Questions  & Question: who won the 2017 ncaa mens basketball tournament                                                                                                                              & North Carolina                                                      \\ \hline
Evidence Inference & Question: With respect to Quality of life, characterize the reported difference between patients receiving good motivation/capability and those receiving inadequate motivation/capability.Choose between 'significantly decreased', 'no significant difference', and 'significantly increased'.
                                                      & no significant difference                                           \\ \hline
Wice               & Claim: Having over 3,000 animals of nearly 400 different species, the zoo has slowly increased its visitors and now ranks as the number one outdoor tourist attraction in the state. \newline Additional Info: The Sedgwick County Zoo is an AZA-accredited wildlife park and major attraction in Wichita, Kansas.  Founded in 1971 with the help of the Sedgwick County Zoological Society, the zoo has quickly become recognized both nationally and internationally for its support of conservation programs and successful breeding of rare and endangered species. & Partially Supported                                                 \\ \hline
ContractNLI        & Claim: Receiving Party shall not reverse engineer any objects which embody Disclosing Party's Confidential Information.                                                              & not mentioned                                                       \\ \hline
GovReport          &  Question: Write a one-page summary of the document. Structure your summary according to the following questions: 1. Why GAO Did This Study 2. What GAO Found 3. What GAO Recommends                                                     & \{summary\}     \\
\bottomrule
\end{tabularx}
\caption{Dataset examples}
\label{tab:dataset_examples}
\end{table*}

\begin{table*}[]
\small
\centering
\begin{tabularx}{\textwidth}{l|X}
\toprule
QASPER             & You are given a Scientific Article and a Question. Answer the Question as concisely as you can, using a single phrase or sentence. If the Question cannot be answered based on the information in the Article, answer "unanswerable". If the question is a yes/no question, your answer should be "yes", "no", or "unanswerable". Do not provide any explanation. (If the question can be answered, provide one or several evidence paragraphs that can be used to verify the answer. Give as few paragraphs as possible.) \\ \hline
Natural Questions  & You are given a Wikipedia page and a question about that page. Answer the question as concisely as you can, using at most five (5) words. If the question cannot be answered based on the information in the article, write "unanswerable". If the question is a yes/no question, answer "yes", "no", or "unanswerable". Do not provide any explanation. (If the question can be answered, provide one evidence paragraph that can be used to verify the answer.)                                                          \\ \hline
Evidence Inference & You are given a clinical study report and a question. Assess the effect of a treatment on a clinical outcome, compared to a control treatment. The options are "significantly increased", "significantly decreased" and "no significant difference". Do not provide any explanation. (Provide one or several evidence paragraphs that can be used to verify the answer. Give as few paragraphs as possible.)                                                                                                               \\ \hline
Wice               & You are given a document and a claim. Evaluate if the claim is supported by the document. You can choose between “supported”, “partially supported” and “not supported”. Do not add any explanation. (If you answer “supported” or “partially supported”, provide the evidence sentences from the document that can be used to verify the answer. Give as few sentences as possible.)                                                                                                                                      \\ \hline
ContractNLI        & You are given a non disclosure agreement contract and a statement. Determine the relation between the contract and the statement. You can choose between “entailment”, “contradiction” and “not mentioned”. Do not add any explanation. (If you answer “entailment” or “contradiction”, provide the evidence paragraphs from the contract that can be used to verify the answer. Give as few paragraphs as possible.)                                                                                                      \\ \hline
GovReport          & You are given a government report document. Write a one page summary of the document. (Each sentence in your summary should reference the source paragraphs from the document that can be used to verify the summary sentence.)      \\
\bottomrule
\end{tabularx}
\caption{Task explanations for the datasets in \benchmarkname{}. Text in parentheses at the end was only shown when \texttt{citation} approaches were used.}
\label{tab:task_explanations}
\end{table*}
\begin{table*}[]
\centering
\small
\begin{tabularx}{\textwidth}{l|X}
\toprule
Single statement & Your reply must have the following format:  \newline 
"\textless{}answer\textgreater{} {[}X{]} {[}Y{]}" \newline
In your reply, replace \textless{}answer\textgreater{} with your solution to the task. Your reply must be followed by the ids of the relevant segments from the document.      \\                                                                                                                                                        \hline
Multi statement  & Your reply must have the following format: \newline
"\textless{}answer\_sentence\_1\textgreater {[}X{]} {[}Y{]} \textless{}answer\_sentence\_2\textgreater {[}Z{]}..." \newline
In your reply, replace \textless{}answer\_sentence\_1\textgreater{} with your first sentence, \textless{}answer\_sentence\_2\textgreater{} with your second sentence, and so forth. Each sentence must be followed by the ids of the segments relevant to the sentence. \\ 
\bottomrule
\end{tabularx}
\caption{Format explanations. Multi statement was used for GovReport, Single Statement for all other datasets.}
\label{tab:format_explanations}
\end{table*}
\begin{table}[]
\centering
\small
\begin{tabularx}{7cm}{l|X}
\toprule
QSP             & Scientific Article: \{document\} {[}End of Document{]}\newline Question: \{question\}                            \\ \hline
NQ  & Document: \{document\} {[}End of Document{]}\newline Question: \{question\}                                    \\ \hline
EI & Document: \{document\} {[}End of Document{]}\newline Question: \{question\}                                      \\ \hline
WIC               & Document: \{document\} {[}End of Document{]}\newline Claim: \{statement\}\newline Additional Info: \{additional\_info\} \\ \hline
CNLI        & Contract: \{document\} {[}End of Document{]}\newline Statement: \{statement\}                                  \\ \hline
GR          & Document: \{document\} {[}End of Document{]}\newline \{question\}  \\
\bottomrule
\end{tabularx}
\caption{Formatting of instance specific input. See Table~\ref{tab:dataset_examples} for examples of task specific inputs.}
\label{tab:instance_specific_input}
\end{table}
\begin{table*}[]
\centering
\small
\begin{tabular}{l|cccc|ccc}
\toprule
                        & \multicolumn{4}{c}{QSP} &  \multicolumn{3}{c}{GR}    \\
                        & AF1    & ATT & UF1 & Avg   & RL        & ATT & Avg   \\
\midrule
Inst before and after   & 42     & 75  & 14  & 43.77 & 22        & 43  & 32.60 \\
Inst before             & 47     & 70  & 33  & 50.31 & 25        & 10  & 17.62 \\
Inst after              & 40     & 60  & 20  & 40.16 & 21        & 14  & 17.85 \\
Inst before, 1 example  & 53     & 67  & 43  & 54.21 & 26        & 36  & 30.97 \\
Inst before, 2 examples & 50     & 59  & 42  & 50.40 & 26        & 44  & 35.11 \\
Inst before, 3 examples & 50     & 66  & 67  & \textbf{60.86} & 28        & 54  & \textbf{41.08} \\
\bottomrule
\end{tabular}
\caption{Prompt optimization results on GPT-3.5 under the \texttt{citation} approach. Inst before / after refers to the position of the instruction being before / after the task specific input. For complete instructions, see \ref{sec:appendix/prompts} Avg: Average of scores for one task.}
\label{tab:prompt_optimization}
\end{table*}

\begin{table*}[]
\small
\centering
\begin{tabularx}{\textwidth}{r|X}
\toprule
Task Explanation           & You are given a Scientific Article and a Question. Answer the Question as concisely as you can, using a single phrase or sentence. If the Question cannot be answered based on the information in the Article, answer "unanswerable". If the question is a yes/no question, your answer should be "yes", "no", or "unanswerable". Do not provide any explanation. (If the question can be answered, provide one or several evidence paragraphs that can be used to verify the answer. Give as few paragraphs as possible.) \\
Format explanation  & Your reply must have the following format:  \newline 
"\textless{}answer\textgreater{} {[}X{]} {[}Y{]}" \newline
In your reply, replace \textless{}answer\textgreater{} with your solution to the task. Your reply must be followed by the ids of the relevant segments from the document.                                                           \\ 
Example 1 &    Scientific Article:  Automated Hate Speech Detection and the Problem of Offensive Language \newline
Abstract \{omitted\} {[}End of Document{]}\newline Question: What type of model do they train?\                                                                                                            \\ 
Example 2               & \{omitted\} \\
Example 3       & \{omitted\}                                                                                                      \\
Instance specific input          & Scientific Article: Combining Thesaurus Knowledge and Probabilistic Topic Models \newline
Abstract \{omitted\} {[}End of Document{]}\newline Question: Which domains do they explore?\      \\
\bottomrule
\end{tabularx}
\caption{Example of the final prompt format used for \texttt{citation} on QASPER.}
\label{tab:full_prompt_example}
\end{table*}

\begin{table*}[]
\small
\centering
\begin{tabularx}{\textwidth}{l|l|X}
\toprule
 & Input & Claim \\
\midrule
QASPER & Question, response & ``The answer to the question 'Which domains do they explore?' is 'news articles related to Islam and articles discussing Islam basics''' \\ \hline
Natural Questions & Question, response & ``The answer to the question 'who won the 2017 ncaa mens basketball tournament?' is 'North Carolina''' \\ \hline
"Evidence Inference & Question, response & ``There was no significant difference between the effect of good motivation/capability and the effect of inadequate motivation/capability on quality of life.'' \newline
``Good motivation/capability [significantly increased/significantly decreased] quality of life compared to inadequate motivation/capability''' \\ \hline
"Wice & Claim, response & ``Having over 3,000 animals of nearly 400 different species, the zoo has slowly increased its visitors and now ranks as the number one outdoor tourist attraction in the state.'' \newline
``The claim 'Having over 3,000 animals of nearly 400 different species, the zoo has slowly increased its visitors and now ranks as the number one outdoor tourist attraction in the state' is partially supported.''\\ \hline
"ContractNLI & Claim, response & ``Receiving Party shall not reverse engineer any objects which embody Disclosing Party’s Confidential Information.'' \newline
``Receiving Party may reverse engineer any objects which embody Disclosing Party’s Confidential Information.'''\\ \hline
GovReport & Response statement & ``Improper payments in Medicaid increased from \$29.1 billion in fiscal year 2015 to \$36.7 billion in fiscal year 2017.'' \\
\bottomrule
\end{tabularx}
\caption{Examples for claims constructed for attributability evaluation.}
\label{tab:claim_examples}
\end{table*}
\begin{table*}[]
\small
\centering
\begin{tabularx}{\textwidth}{r|X}
\toprule
TRUE & premise: \{evidence\} hypothesis: \{claim\} \\ \hline
Attrscore & Below is an instruction that describes a task, paired with an input that provides further context. Write a response that appropriately completes the request.\newline
\#\#\# Instruction:\newline
Verify whether a given reference can support the claim. Options: ""Attributable, Extrapolatory or Contradictory.\newline
\#\#\# Input:\newline
Claim: \{claim\}\newline
Reference: \{evidence\}\newline
\#\#\# Response: \\ \hline
Minicheck & predict: \{evidence\}\textless/s\textgreater\{claim\} \\
\bottomrule
\end{tabularx}
\caption{Prompts for attributability evaluation models based on the respective publications.}
\label{tab:attributability_evaluation_prompts}
\end{table*}

\begin{table}[]
    \centering
    \small
    \begin{tabularx}{7cm}{X|l|X}
        \toprule
         & \#Params & Reference \\
         \midrule
         Gemma-7b-it & 7B & \citet{gemmateam2024gemma} \\ \hline
GritLM-7B & 7B & \citet{muennighoff2024generative} \\ \hline
Mistral-7B-Instruct-v0.2 & 7B & \citet{jiang2023mistral} \\ \hline
Vicuna-7B-v1.5-16K & 7B & \citet{vicuna2023} \\ \hline
Flan-T5-XL/XXL* & 3B/11B & \citet{pmlr-v202-longpre23a} \\ \hline
LongChat-7B-v1.5-32K & 7B & \citet{longchat2023} \\ \hline
Llama3-8B-Instruct & 8B & See \footnote{\url{https://llama.meta.com/llama3/}} \\
         \bottomrule
    \end{tabularx}
    \caption{Open source models considered in selection experiments. *: Flan-T5-XL was used in fine-tuning, Flan-T5-XXL was used in prompting experiments.}
    \label{tab:candidate_models}
\end{table}
\begin{table}
\small
\centering 
\begin{tabular}{l|ccc|c}
\toprule
 \multicolumn{1}{c|}{} & \multicolumn{1}{c}{AF1} & \multicolumn{1}{c}{ATT} & \multicolumn{1}{c|}{UF1} & \multicolumn{1}{c}{Avg}   \\
\midrule
\textbf{\texttt{Gemma}} & 0.41 & 0.28 & 0.25 & 0.27   \\
\textbf{\texttt{GritLM}} & 0.42 & 0.30 & 0.22 & 0.26    \\
\textbf{\texttt{Mistral}} & \textbf{0.46} & 0.34 & \textbf{0.26} & 0.30  \\
\textbf{\texttt{Vicuna}} & 0.42 & 0.31 & 0.24 & 0.28  \\
\textbf{\texttt{Flan-T5-XL}} & 0.45 & \textbf{0.61} & 0.22 & \textbf{0.42}   \\
\textbf{\texttt{LongChat}} & 0.42 & 0.29	& 0.24	& 0.26 \\
\textbf{\texttt{Llama3}} & 0.44 & 0.37 & 0.25 &	0.31  \\
\bottomrule
\end{tabular}
\caption{results of Fine-tuning Model Selection}
\label{tab:finetune_model_selection}

\end{table}

\begin{table*}
\small
\centering 
\begin{tabular}{l||ccc|cc||c}
\toprule
 & \multicolumn{3}{c|}{QASPER} & \multicolumn{2}{c|}{GovReport} & \multicolumn{1}{c}{Avg}   \\
Model & Answer F1 & Attr & Unans F1 & R-L & Attr & \textbf{}  \\
\midrule
\texttt{\textbf{Gemma-7b-it}}              & 22          & 2           & 49          & 18          & 0           & 17          \\
\texttt{\textbf{GritLM-7B}}                & 26          & 6           & 48          & 20          & 0           & 18          \\
\texttt{\textbf{Mistral-7B-Instruct-v0.2}} & \textbf{31} & \textbf{32} & 48          & \textbf{24} & 0           & \textbf{25} \\
\texttt{\textbf{Vicuna-7B-v1.5-16K}}       & 21          & 12          & 49          & 21          & 2           & 19          \\
\texttt{\textbf{Flan-T5-XXL}}              & 26          & 19          & \textbf{66} & 12          & 1           & 22          \\
\texttt{\textbf{LongChat-7B-v1.5-32K}}     & 21          & 18          & 48          & 22          & \textbf{13} & 24          \\
\texttt{\textbf{Llama3-8B}}                & 17          & 17          & 47          & 22          & 0           & 19         \\
\bottomrule
\end{tabular}
\caption{Results of model selection for prompted open source models.}
\label{tab:prompt_model_selection}

\end{table*}


\begin{table*}
\small
\centering
\begin{tabular}{l|cccc|cccc}
\toprule
& \multicolumn{4}{c|}{\textbf{QASPER}} 
& \multicolumn{4}{c}{\textbf{Natural Questions}}                    \\ 
& AF1 & ATT & UF1 & Avg  & AF1 & ATT  & UF1 & Avg   \\
\midrule
BM25 & 32 & 60 & 42 & 44.49  & 44	&33 &	59	& 45.17     \\
SBERT & 29 & 69 & 39	& 45.56  & 47	 & \textbf{53}	&60	&\textbf{53.36} \\
Contriever & 37 &	75	& 50	& 54.20 & 47	& 51 &	61	& 52.89    \\
Dragon & \textbf{39}	& \textbf{79}	&\textbf{52}	&\textbf{56.65} & \textbf{48}	& 48	&\textbf{63}	& 52.93 \\ 
GTR & 36	&70 &	46	&50.54 & 46	& 48 &	61	& 51.47\\
\bottomrule
\toprule
 & \multicolumn{4}{c|}{\textbf{Evidence Inference}} & \multicolumn{4}{c}{\textbf{Wice}} \\ 

 & CF1 & EF1 & Avg & -   & CF1 & EF1 & Avg & -  \\
 \midrule
BM25 & 77	& 16	& 46.46 & -  & 29	& 36 &	32.71 & -   \\
SBERT & 78 &23 &	50.51 & -   & 33 &	43 & \textbf{42.50} & -   \\
Contriever & \textbf{83}	& 29	& \textbf{55.71} & -  & 26	& 42 &	33.87 & -  \\
Dragon & 78	& \textbf{33} &55.23 & -  & 31 &	41 &	36.29 & -    \\ 
GTR & 78	& \textbf{33} &	55.58 & -  & \textbf{33}	& \textbf{43}	&38.20 & -   \\ 
\bottomrule
\toprule
 & \multicolumn{4}{c|}{\textbf{Contract NLI}} & \multicolumn{4}{c}{\textbf{Govreport}} \\ 
 & CF1 & EF1 & Avg & - & RL  & ATT & Avg \\
\midrule
BM25 & 43	& 34	&38.35 & -  & 27	& 33	&29.96 & -  \\
SBERT & 42 &35	&38.21 & -   & \textbf{28}	& 36 &	\textbf{31.79} & -   \\
Contriever & 38	& 37 & 36.80 & -  & 24 &	30 &	26.92  & - \\
Dragon & 39 & 37	& 37.64 & -  & 21 & 19 & 20.16 & -   \\ 
GTR & \textbf{44}	& \textbf{39}	&\textbf{41.39} & -  & 23	& \textbf{37} &	30.00  & -    \\ 

\bottomrule
\end{tabular}
\caption{Retriever selection for \texttt{retrieve-then-read}. Retrievers were combined with GPT-3.5 and were evaluated on 100 dev instances per dataset. The retriever that resulted in the best average score was used in all further experiments for the respective combination of \texttt{retrieve-then-read} and task.}
\label{tab:rtr_retriever_selection}
\end{table*}

\begin{table*}
\small
\centering
\begin{tabular}{l|cccc|cccc}
\toprule
& \multicolumn{4}{c|}{\textbf{QASPER}} 
& \multicolumn{4}{c}{\textbf{Natural Questions}}                    \\ 
& AF1 & ATT & UF1 & Avg  & AF1 & ATT  & UF1 & Avg   \\
\midrule
BM25 & 45 & 67 & \textbf{63} & 58.33 & \textbf{42} & 36 & \textbf{59} & 45.69 \\
SBERT & 40 & 65 & 51 & 51.79 & 41 & 39 & 55 & 45.13 \\
Contriever & 48 & 72 & 65 & \textbf{61.64} & 40 & \textbf{42} & 55 & 45.83 \\
Dragon & \textbf{49} & \textbf{73} & 56 & 59.20 & \textbf{42} & \textbf{42} & 58 & \textbf{46.95} \\
GTR & 47 & 71 & \textbf{63} & 60.17 & \textbf{42} & 41 & 58 & 46.63 \\
\bottomrule
\toprule
 & \multicolumn{4}{c|}{\textbf{Evidence Inference}} & \multicolumn{4}{c}{\textbf{Wice}} \\ 

 & CF1 & EF1 & Avg & -   & CF1 & EF1 & Avg & -  \\
 \midrule
BM25 & 83 & 49 & 66.12 & - & \textbf{44} & 61 & \textbf{52.38} & -  \\
SBERT & 82 & 50 & 65.75 & - & 40 & \textbf{65} & 52.13 & - \\
Contriever & 83 & 57 & 70.42 & - & 36 & 60 & 48.14 & - \\
Dragon & \textbf{87} & 54 & 70.51 & - & 37 & 64 & 50.49 & - \\
GTR & 86 & \textbf{62} & \textbf{73.66} & - & 37 & 62 & 49.75 & - \\
\bottomrule
\toprule
 & \multicolumn{4}{c|}{\textbf{Contract NLI}} & \multicolumn{4}{c}{\textbf{Govreport}} \\ 
 & CF1 & EF1 & Avg & - & RL  & ATT & Avg \\
\midrule
BM25 & 42 & 49 & 45.25 & - & 26 & 51 & 38.29 & - \\
SBERT & 44 & 56 & 50.11 & - & \textbf{27} & \textbf{60} & \textbf{43.38} & - \\
Contriever & 45 & \textbf{58} & 51.42 & - & 23 & 45 & 33.93 & - \\
Dragon & \textbf{52} & 55 & \textbf{53.29} & - & 22 & 43 & 32.12 & - \\
GTR & 46 & 54 & 49.94 & - & 23 & 51 & 37.23 & - \\
\bottomrule
\end{tabular}
\caption{Retriever selection for \texttt{reduced} approaches. Retrievers were combined with GPT-3.5-\texttt{reduced-citation} and were evaluated on 100 dev instances per dataset. The retriever the resulted in the best average score was used in all further experiments for the respective combination of \texttt{reduced-citation} / \texttt{reduced-post-hoc} and task.}
\label{tab:reduce_retriever}
\end{table*}

\begin{table*}
\small
\centering
\begin{tabular}{l|cccc|cccc}
\toprule
& \multicolumn{4}{c|}{\textbf{QASPER}} 
& \multicolumn{4}{c}{\textbf{Natural Questions}}                    \\ 
& AF1 & ATT & UF1 & Avg  & AF1 & ATT  & UF1 & Avg   \\
\midrule
BM25 & 52 & 65 & 73 & 63.25 & 41 & 40 & 57 & 45.92 \\
SBERT & 52 & 55 & 73 & 64.12 & 41 & 50 & 57 & 53.35 \\
Contriever & 52 & 63 & 73 & 68.42 & 41 & \textbf{54} & 57 & \textbf{55.28} \\
Dragon & 52 & \textbf{69} & 73 & \textbf{71.11} & 41 & \textbf{54} & 57 & \textbf{55.28} \\
GTR & 52 & 59 & 73 & 66.27 & 41 & \textbf{54} & 57 &\textbf{55.28} \\
\bottomrule
\toprule
 & \multicolumn{4}{c|}{\textbf{Evidence Inference}} & \multicolumn{4}{c}{\textbf{Wice}} \\ 

 & CF1 & EF1 & Avg & -   & CF1 & EF1 & Avg & -  \\
 \midrule
BM25 & 86 & 25 & 55.60 & - & 86 & 25 & 55.60 & - \\
SBERT & 86 & 20 & 52.94 & - & 86 & 20 & 52.94 & - \\
Contriever & 86 & 27 & 56.60 & - & 86 & 27 & 56.60 & - \\
Dragon & 86 & \textbf{28} &\textbf{ 57.27} & - & 86 & \textbf{28} & \textbf{57.27} & - \\
GTR & 86 & 24 & 55.27 & - & 86 & 24 & 55.27 & - \\
\bottomrule
\toprule
 & \multicolumn{4}{c|}{\textbf{Contract NLI}} & \multicolumn{4}{c}{\textbf{Govreport}} \\ 
 & CF1 & EF1 & Avg & - & RL  & ATT & Avg \\
\midrule
BM25 & 46 & 36 & 41.14 & - & 28 & \textbf{73} & \textbf{50.55} & - \\
SBERT & 46 & 36 & 41.24 & - & 28 & 60 & 44.14 & - \\
Contriever & 46 & 39 & 42.69 & - & 28 & 65 & 46.63 & - \\
Dragon & 46 & 40 & 43.16 & - & 28 & 68 & 48.22 & - \\
GTR & 46 & \textbf{41} & \textbf{43.57} & - & 28 & 61 & 44.61 & - \\
\bottomrule
\end{tabular}
\caption{Retriever selection for \texttt{post-hoc} approaches. Retrievers were combined with GPT-3.5-\texttt{post-hoc} and were evaluated on 100 dev instances per dataset. The retriever the resulted in the best average score was used in all further experiments for the respective combination of \texttt{post-hoc} and task.}
\label{tab:post_hoc_retriever_selection}
\end{table*}

\begin{table*}[]
\small
\centering
\begin{tabularx}{\textwidth}{l|l|X}
\toprule
 & Input & Query \\
\midrule
QASPER & Question, response & ``Which domains do they explore? news articles related to Islam and articles discussing Islam basics'' \\ \hline
Natural Questions & Question, response & ``who won the 2017 ncaa mens basketball tournament? North Carolina'' \\ \hline
Evidence Inference & Question, response & ``There was no significant difference between the effect of good motivation/capability and the effect of inadequate motivation/capability on quality of life.'' \newline
``Good motivation/capability [significantly increased/significantly decreased] quality of life compared to inadequate motivation/capability'' \\ \hline
Wice & Claim & ``Having over 3,000 animals of nearly 400 different species, the zoo has slowly increased its visitors and now ranks as the number one outdoor tourist attraction in the state'' \\ \hline
ContractNLI & Claim, response & ``Receiving Party shall not reverse engineer any objects which embody Disclosing Party’s Confidential Information.'' \newline
``Receiving Party may reverse engineer any objects which embody Disclosing Party’s Confidential Information.'' \\ \hline
GovReport & Response statement & ``Improper payments in Medicaid increased from \$29.1 billion in fiscal year 2015 to \$36.7 billion in fiscal year 2017.'' \\\hline
\bottomrule
\end{tabularx}
\caption{Examples for queries for \texttt{post-hoc} evidence retrieval.}
\label{tab:post_hoc_query_examples}
\end{table*}
\begin{table*}[]
\small
\centering
\begin{tabularx}{\textwidth}{l|l|X}
\toprule
 & Input & Query \\
\midrule
QASPER & Question & ``Which domains do they explore?'' \\ \hline
Natural Questions & Question & ``who won the 2017 ncaa mens basketball tournament?'' \\ \hline
Evidence Inference & Question & ``With respect to Quality of life, characterize the reported difference between patients receiving good motivation/capability and those receiving inadequate motivation/capability. Choose between ’significantly decreased’, ’no significant difference’, and  ’significantly increased’.'' \\ \hline
Wice & Claim & ``Having over 3,000 animals of nearly 400 different species, the zoo has slowly increased its visitors and now ranks as the number one outdoor tourist attraction in the state'' \\ \hline
ContractNLI & Claim & ``Receiving Party shall not reverse engineer any objects which embody Disclosing Party’s Confidential Information.'' \\ \hline
GovReport & Paragraph & ``Medicaid has been on our high-risk list since 2003, in part, because of  concerns about the adequacy of fiscal oversight and the program’s  improper payments—including payments made...'' \\
\bottomrule
\end{tabularx}
\caption{Examples for queries for \texttt{retrieve-then-read} and \texttt{reduced} retrieval.}
\label{tab:retrieve-then-read_query_examples}
\end{table*}

\begin{figure*}
    \centering
    \includegraphics[page=1, width=\textwidth]{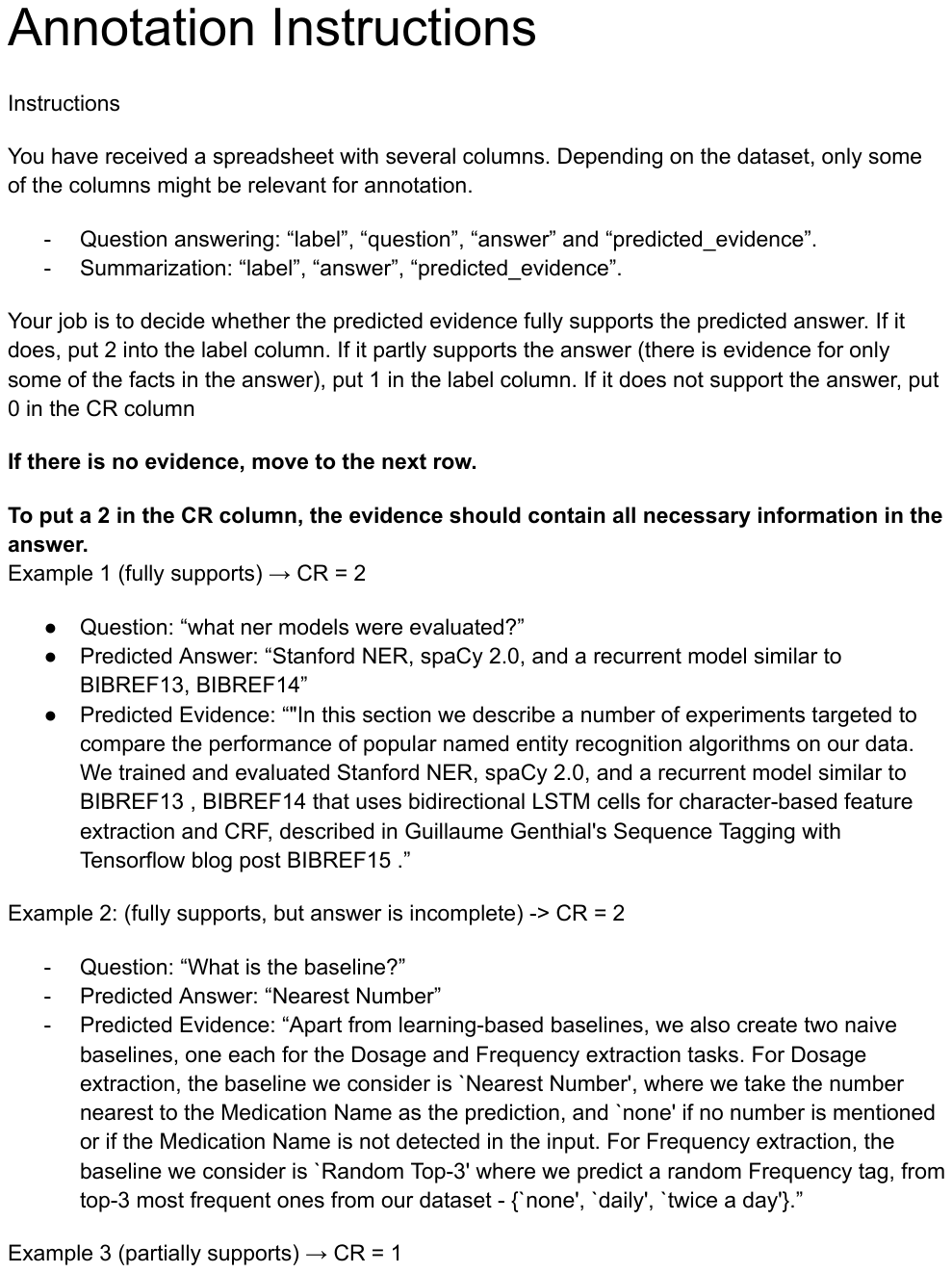}
\end{figure*}

\begin{table}[]
\small
\begin{tabular}{l}
\toprule
unanswerable              \\
n/a                       \\
i don't know              \\
idk                       \\
not known                 \\
answer not in context     \\
unknown                   \\
no answer                 \\
it is unknown             \\
the answer is unknown     \\
unavailable               \\
not clear                 \\
i cannot provide          \\
i cannot directly provide \\
i cannot answer           \\
i cannot display          \\
not clear                 \\
not available             \\
not readily available    \\
\bottomrule
\end{tabular}
\caption{Simple keyphrases to detect abstaining. See §\ref{sec:appendix/technical_details} for details on how these were used.}
\label{tab:unanswerable_keywords}
\end{table}
\begin{table}[]
\small
\begin{tabular}{l|l}
\toprule
Verbs   & Adverbs      \\
\midrule
provide & explicitly   \\
mention & specifically \\
state   & directly     \\
specify & clearly      \\
define  &              \\
report  &              \\
name    &              \\
offer   &             \\
\bottomrule
\end{tabular}
\caption{Verbs and adverbs that were combined to phrases to detect abstaining. See §\ref{sec:appendix/technical_details} for details on how these were used.}
\label{tab:unanswerable_phrases}
\end{table}

\end{document}